\documentclass[journal,twoside]{IEEEtran}
\ifCLASSINFOpdf
\else
\fi
\usepackage{graphicx} 
\graphicspath{{./pdf/}} 
\DeclareGraphicsExtensions{.pdf,jpg}

\usepackage[export]{adjustbox} 

\usepackage[skip=1pt,labelformat=empty]{subcaption} 

\usepackage[cmex10]{amsmath} 

\usepackage[short]{optidef} 

\usepackage[ruled,linesnumbered]{algorithm2e} 

\usepackage{caption} 

\usepackage{bm} 

\usepackage{epsfig} 

\usepackage{times} 

\usepackage{color} 

\usepackage{amssymb}  
\usepackage{setspace}

\usepackage{amsthm}

\newcommand{\mytilde}{\raise.17ex\hbox{$\scriptstyle\mathtt{‌​\sim}$}}

\usepackage{xparse,soul}

\usepackage{multirow}
\usepackage{diagbox}

\usepackage{cite}

\usepackage{soul}
\soulregister\cite7 
\soulregister\ref7

\usepackage[switch]{lineno}

\begin{document}
%
\title{FBG-Based Online Learning and 3-D Shape Control of Unmodeled Continuum and Soft Robots in Unstructured Environments}
%
%
%

\author{Yiang~Lu, Wei~Chen, Bo~Lu, Jianshu~Zhou, Zhi~Chen, Qi~Dou, and Yun-Hui~Liu
\thanks{This work was supported in part by Shenzhen Portion of Shenzhen-Hong Kong Science and Technology Innovation Cooperation Zone under HZQB-KCZYB-20200089, in part of the HK RGC under T42-409/18-R and 14202918, in part by the Hong Kong Centre for Logistics Robotics, in part by the Multi-Scale Medical Robotics Centre, InnoHK, and in part by the VC Fund 4930745 of the CUHK T Stone Robotics Institute. \textit{(Corresponding author: Jianshu Zhou and Bo Lu.)}}
\thanks{Y. Lu, W. Chen, J. Zhou, Z. Chen, and Y.-H. Liu are with T Stone Robotics Institute, Department of Mechanical and Automation Engineering, The Chinese University of Hong Kong, Hong Kong.}
\thanks{Q. Dou is with the Department of Computer Science and Engineering, The Chinese University of Hong Kong, Hong Kong.}
\thanks{J. Zhou and Y.-H. Liu are also with the Hong Kong Center for Logistics Robotics, Hong Kong.}
\thanks{B. Lu is with the Robotics and Microsystems Center, School of Mechanical and Electric Engineering, Soochow University, Suzhou, Jiangsu, China}
}

%
%

\markboth{}%
{}
%



\maketitle

\begin{abstract}
In this paper, we present a novel and generic data-driven method to servo-control the 3-D shape of continuum and soft robots embedded with fiber Bragg grating (FBG) sensors.
Developments of 3-D shape perception and control technologies are crucial for continuum robots to perform the tasks autonomously in surgical interventions.
However, owing to the nonlinear properties of continuum robots, one main difficulty lies in the modeling of them, especially for soft robots with variable stiffness.
To address this problem, we propose a versatile learning-based adaptive controller by leveraging FBG shape feedback that can online estimate the unknown model of continuum robot against unexpected disturbances and exhibit an adaptive behavior to the unmodeled system without priori data exploration.
Based on a new composite adaptation algorithm, the asymptotic convergences of the closed-loop system with learning parameters have been proven by Lyapunov theory.
To validate the proposed method, we present a comprehensive experimental study by using two continuum robots both integrated with multi-core FBGs, including a robotic-assisted colonoscope and multi-section extensible soft manipulators.
The results demonstrate the feasibility, adaptability, and superiority of our controller in various unstructured environments as well as phantom experiments.
\end{abstract}


%
\IEEEpeerreviewmaketitle

\section{Introduction}
%
%
%
%
Continuum and soft robots are composed of compliant materials and flexible structures, endowing them with high dexterity and compliance for a wide variety of application scenarios, such as industrial fields, human-robot interactions, and medical interventions.
In view of the advantages of continuum robots in terms of manipulating accessibility, intraoperative maneuverability, and inherent safety, they possess great potentials to be deployed in the treatment of various types of minimally invasive surgery (MIS) \cite{webster2010design,burgner2015continuum}.
However, the interference of nonlinear structural characteristics and unknown environmental disturbances would negatively affect the priori modeling of continuum robots, hence accurate 3-D shape sensing and robust control are considered as major challenges for their efficient and safe implementations in unstructured surgical scenarios (as illustrated in Figure 1), 
especially for soft manipulators 
with heterogeneous properties \cite{shi2016shape, george2018control,wang2021survey,sefati2021dexterous,fang2021soft,dong2022shape}.

Considering that continuum robot has an infinite number of degree-of-freedoms (DOFs) in configuration space \cite{burgner2015continuum, george2018control}, real-time perception of its 3-D shape is essential to advance control strategies.
To date, numerous types of sensing modalities have been exploited for shape sensing and position estimation of continuum robots in MIS, such as mechanics-based methods, intraoperative imaging modalities, magnetic resonance imaging, and electromagnetic tracking \cite{shi2016shape}.
However, these methods suffer from their limitations concerning acquisition frequency, sensing accuracy, application versatility, and clinical compatibility.
Recently, fiber Bragg grating (FBG) sensors have shown great potentials in shape sensing of continuum robots and flexible instruments \cite{chitalia2020towards, sefati2021dexterous, dong2022shape} due to their lightweight, high compatibility, and high feedback frequency without requiring line-of-sight.
Considering severe signal noises of FBG measurements in unstructured environments, the accumulative error along the longitudinal direction could defect the accuracy of the sensing result.
Aiming to solve this problem, some sensor fusion using Kalman filter (KF) \cite{denasi2018observer, alambeigi2019scade, donder2021kalman} and learning-based approaches \cite{wang2020eye, sefati2020data} have been developed.
A model-based filtering algorithm for 3-D shape sensing of flexible endoscope using multi-core FBGs is recently reported to be accurate and robust against the sensory noises and unexpected perturbations \cite{lu2021robust}.

Employing task-space position/shape with encoders information as feedback, various control methodologies of continuum robots have been proposed for distal endpoint positioning and shape servoing \cite{xu2019adaptive, alambeigi2020versatile}.
According to modeling approaches, control strategies can be mainly classified into model-based and model-free methods.
Model-based methods require a priori accurate modeling of the continuum robot, including kinematics/statics and dynamics models \cite{huang2022kinematic} based on certain assumptions and theories, e.g., constant curvature \cite{webster2010design, wang2016visual}, beam theory \cite{camarillo2009configuration},
and Cosserat rod theory \cite{rucker2011statics, campisano2021closed}.
In model-based methods, the modeling of continuum robots should be identified before implementation which would result in unknown deviations in the presence of backlash, friction, and hysteresis.
In addition, their strong assumptions might not be validated with unexpected disturbances.
Model-free control strategies do not require a priori knowledge about the model of continuum robot and its modeling uncertainties, which are instead acquired by estimation methods.
They are applicable to the highly nonlinear or nonuniform systems that can be affected by unknown perturbations, making priori models invalid \cite{george2018control}.
Yip and Camarillo \cite{yip2014model,yip2016model} proposed optimal control methods with optimization-based online estimations of the kinematics Jacobian and stiffness 
of a continuum robot. 
Alambeigi \textit{et al.} \cite{alambeigi2019scade, alambeigi2020versatile} presented a model-independent KF to estimate the deformation of an unmodelled continuum robot by fusing eye-to-hand 
vision and FBG signals; lately, they introduced an optimization-based control framework for visual servoing of the robot with online model update using Broyden's method.
Bruder \textit{et al.} \cite{bruder2020data} introduced a data-driven model predictive control (MPC) strategy
for soft robots by incorporating a Koopman-based system identification procedure.

With the advancements of artificial intelligence, control strategies employing machine learning, which can identify and learn the model/motion-based policies of continuum robots, have also gained great attention \cite{wang2021survey}.
Amongst, neural networks (NNs) are commonly utilized for modeling of continuum robots (e.g., inverse kinematics) \cite{giorelli2015neural, george2017learning, bern2020soft, fang2021soft, dong2022shape}.
Gaussian process regression has been also investigated as alternative approach to approximate models of soft robots \cite{fang2019vision, hamaya2021design}.
Thuruthel \textit{et al.} \cite{thuruthel2018model} implemented reinforcement learning (RL) for soft robot control through learning control actions from recurrent NN-based dynamics model.
Model-free RL algorithms are also developed to control soft robot by using simulation data instead of analytical modeling \cite{you2017model, jiang2021hierarchical}.
Although these approaches alleviate the need for accurate modeling of continuum robots, large amounts of training data and offline identification are heavily required, hampering their practical deployments \cite{wang2021survey}.
Due to probabilistic properties of these learning-based approaches, another major concern is the difficulty of deriving stability conditions and even asymptotic convergences of control error and learning parameters \cite{tang2021model}, which may raise safety problems in unstructured contexts.

\begin{figure}[t]
  \centering
  \includegraphics[width=0.9\linewidth]{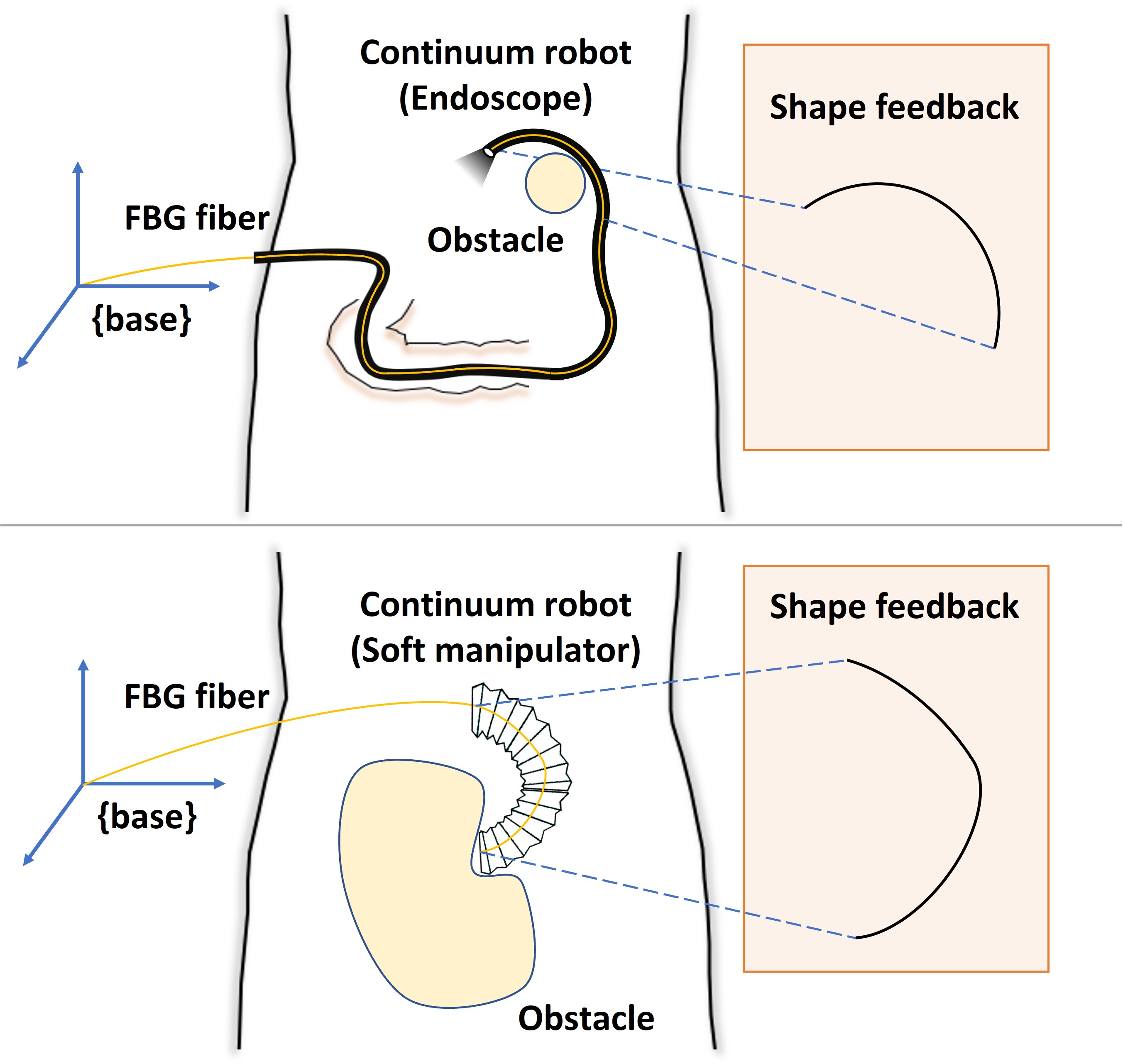}
   \caption{Conceptual representation of 3-D shape sensing and control of continuum robots, e.g., endoscope (up), and soft manipulator (bottom), in unstructured environments.}
  \label{fig:illustration}
\end{figure}

To guarantee convergent performance when using learning-based methods,
Braganza \textit{et al.} \cite{braganza2007neural} utilized a NN to feedforward account for the 
dynamics uncertainty in a soft continuum robot, while maintaining the boundedness of NN output together with the tracking error convergence
under a nonlinear feedback controller.
Tang \textit{et al.} \cite{tang2021model} developed an integration of iterative learning and MPC for soft robots that can online update the model parameters and guarantee the asymptotic stability.
However, control strategies of continuum robots with stability analysis concerning the 
modeling error caused by online 
learning 
have not been reported \cite{slotine1991applied}.
Nevertheless, Melingui \textit{et al.} \cite{melingui2015adaptive} addressed the stability of an adaptive NN controller that learns the kinematics of continuum robots, training an emulator is necessarily required before the control task.
Moreover, 
some existing methodologies 
were only validated with motions in 2-D space by adding in-plane perturbations, while neglecting those out-of-plane ones, and thus without investigations in 3-D shape \cite{alambeigi2020versatile, xu2021visual}.
Overall, there are open problems in 3-D shape control of continuum robots enabling them to perform complex tasks in sensitive environments \cite{george2018control}, especially for learning-based strategies, which are expected to integrate the advantages:
no priori data requirement, stability with convergences of control error and learning parameters, online error compensation, and accurate 
feedback \cite{burgner2015continuum, wang2021survey}.

Towards these limitations, this paper proposes a robust data-driven control framework, which can automatically regulate the 3-D shape of FBG-equipped continuum robot in unstructured environments.
To the best of the authors' knowledge, this is the first work in control with guaranteed stability for unmodeled continuum robots and soft continuum robots using 3-D shape feedback (e.g., FBG sensing), based on the development of a novel NN-based adaptive control strategy for online learning the robot's differential kinematics and simultaneous shape servoing.
The proposed algorithm does not require priori data exploration and can concurrently guarantee the convergent performance of control error and learning parameters, and online error compensation with unknown external conditions as compared with the state-of-the-art methods \cite{wang2021survey}.
The main contributions are summarized as follows:
\begin{enumerate}
    \item Development of a data-driven adaptive controller by incorporating FBG-based shape feedback and a learning-based shape flow predictor, to online update the unknown differential kinematics of the continuum robot without its priori knowledge, and to simultaneously servo-control its 3-D shape.
    \item Consideration of the online modeling (parameters) error, NN approximation error, and unknown environmental conditions when developing learning-based control of continuum robot. Under the proposed controller together with a novel composite adaptation algorithm jointly introducing the shape estimation and flow error of the robot, 
    the asymptotic stability of the closed-loop system and the convergences of the shape control and estimation errors to zeros are proved using Lyapunov method, thus showing the learning parameters convergence.
    \item Design and implementation of a robust and accurate filtering algorithm for 3-D shape sensing of the continuum robot by solely embedding a multi-core FBG fiber, which is independent of external sensors, and maintains high sensing quality against disturbances.
    More configuration-space features computed from the distributed FBG signals can be leveraged for online learning and shape control than only endpoint position.
    \item Experimental validations of the proposed framework using two FBG-equipped continuum robots, including a robotic-assisted colonoscope system and multi-section extensible soft manipulators.
    The results demonstrate the versatility and adaptability of our algorithm in various unstructured environments as well as two phantom experiments for colonoscopy and oropharyngeal sampling.
\end{enumerate}
    

\section{Mathematical Modeling}
\subsection{Problem Statement}
In this work, we can trace the Cartesian positions of the points on the continuum robot from an FBG-based shape sensing system with the compact integration of FBG fiber into the entire robot.
The 3-D shape feature is constructed from the selected points' positions and utilized as the feedback signal for simultaneous model learning and shape servoing.
Given a desired shape of the continuum robot, the objective is to design a velocity controller on the actuation input such that its shape asymptotically approaches to the desired one without any priori and explicit identification of the robot's model.
To address this problem, we first derive the following differential kinematics of continuum robots and shape feature Jacobian matrix, and then design a data-driven adaptive control algorithm with rigorous stability analysis
in the next section.

\subsection{Kinematics of Continuum Robots}
To represent the state of the continuum robot in its configuration space, we first denote $l$ points selected on the robot with their 3-D Cartesian positions $\bm{r}_i = \begin{bmatrix} x_i & y_i & z_i \end{bmatrix}^{\intercal} \in \mathbb{R}^{3}, i \in \{1, 2, ..., l\}$, which are measured from the FBG sensors, and the actuator position input of the robot $\bm{q} = \begin{bmatrix} q_1 & q_1 & \cdots & q_n \end{bmatrix}^{\intercal} \in \mathbb{R}^{n}$.
Then, we define a vector of the Cartesian positions of $l$ selected points $\bm{r} \in \mathbb{R}^{3l}$ as
\begin{equation}
\begin{aligned}
    \bm{r} = 
    \begin{bmatrix}
        \bm{r}_1^{\intercal} & \bm{r}_2^{\intercal} & ... & \bm{r}_l^{\intercal}
    \end{bmatrix}^{\intercal}
\end{aligned}
\end{equation}
The kinematics of the continuum robot mapping the selected points on the robot $\bm{r}$ to the actuation position $\bm{q}$ can be expressed by a smooth nonlinear unknown function as $\bm{r} = \bm{r}(\bm{q}): \mathbb{R}^{n} \mapsto \mathbb{R}^{3l}$, the differential formulation of which as the differential kinematic equation yields
\begin{equation}
\begin{aligned}
    \label{eq:kinematics}
    \dot{\bm{r}} = 
    \bm{J}_r (\bm{q}) \dot{\bm{q}}
\end{aligned}
\end{equation}
where $\bm{J}_r (\bm{q}) \in \mathbb{R}^{3l \times n}$ represents the unknown Jacobian matrix of the continuum robot.
Note that the actuator velocity input of the continuum robot $\dot{\bm{q}}$ can be exactly set from the proposed controller, and its acceleration $\ddot{\bm{q}}$ is bounded \cite{navarro2016automatic}.

\subsection{Shape Feature Jacobian Matrix}
Using the selected points $\bm{r}_i, i \in \{1, 2, ..., l\}$ to describe the overall 3-D shape of the continuum robot, we construct a vector of shape feature as $\bm{x} = \begin{bmatrix} x_1 & x_2 & ... & x_m \end{bmatrix}^{\intercal} \in \mathbb{R}^{m}$ with a smooth function $\bm{x} = \bm{x}(\bm{r}): \mathbb{R}^{3l} \mapsto \mathbb{R}^{m}$, which will be explicitly defined afterwards.
The selected points' velocity $\dot{\bm{r}}$ can be related to the shape feature flow $\dot{\bm{x}} \in \mathbb{R}^{m}$ as
\begin{equation}
\begin{aligned}
    \label{eq:shape}
    \dot{\bm{x}}
    = 
    \bm{J}_x (\bm{r})
    \dot{\bm{r}}
\end{aligned}
\end{equation}
where $\bm{J}_x (\bm{q}) \in \mathbb{R}^{m \times 3l}$ is the shape feature Jacobian matrix.
In a combination of Equations (\ref{eq:kinematics}) and (\ref{eq:shape}), we can relate the actuation velocity $\dot{\bm{q}}$ to the shape flow $\dot{\bm{x}}$ and take an unknown disturbance term $\bm{d} \in \mathbb{R}^{m}$ in consideration representing the system uncertainties and external disturbances in unstructured environments as
\begin{equation}
\begin{aligned}
    \label{eq:modeling}
    \dot{\bm{x}} = 
    \underbrace{
    \bm{J}_x (\bm{r}) \bm{J}_r (\bm{q}) 
    }_{\bm{J}_c (\bm{q})}
    \dot{\bm{q}}
    + \bm{d}
\end{aligned}
\end{equation}
where $\bm{J}_c (\bm{q}) = \bm{J}_x (\bm{r}) \bm{J}_r (\bm{q}) \in \mathbb{R}^{m \times n}$ is denoted as the combined deformation Jacobian matrix of the robot mapping its actuator velocity $\dot{\bm{q}}$ to the shape feedback flow $\dot{\bm{x}}$.
Note that the disturbance $\bm{d}$ and its time derivative $\dot{\bm{d}}$ are bounded, and $\bm{d} \rightarrow \bm{0}$ as $\dot{\bm{q}} \rightarrow \bm{0}$.

\section{Data-Driven Control Algorithm}
A robust data-driven control strategy using FBG shape feedback and adaptive NNs online learning is designed for 3-D shape servoing of the continuum robot.

\subsection{Online Learning of Combined Jacobian}
Considering that the combined Jacobian matrix $\bm{J}_c (\bm{q})$ has unknown and time-varying elements, $\bm{J}_c (\bm{q})$ is hard to be directly estimated using a conventional adaptive mechanism.
We adopt $m$ adaptive NNs $\bm{W}_i \bm{\theta}_i (\bm{q}), i \in \{1, 2, ..., m\}$, employing radial basis functions (RBFs) \cite{lewis2002neuro, li2018vision} to approximate $m$ rows of $\bm{J}_c (\bm{q})$, respectively, with the actuation $\bm{q}$ as input given by 
\begin{equation}
\begin{aligned}
    \label{eq:Jnn}
    \bm{J}_c (\bm{q}) = 
    \begin{bmatrix}
        \bm{W}_1 \bm{\theta}_1 (\bm{q}) & \bm{W}_2 \bm{\theta}_2 (\bm{q}) & \cdots
        & \bm{W}_{m} \bm{\theta}_{m} (\bm{q})
    \end{bmatrix}^{\intercal}
    + \bm{E}_J
\end{aligned}
\end{equation}
where $\bm{W}_i \in \mathbb{R}^{n \times k}, i \in \{1, 2, ..., m\}$, represents the ideal weight matrix of the $i$-th NN with $k$ neurons, $\bm{\theta}_i (\bm{q}) \in \mathbb{R}^{k}, i \in \{1, 2, ..., m\}$,  is the corresponding vector of the activation functions, and the $j$-th neuron, $\theta_{ij}(\bm{q})$, $j \in \{1, 2, ..., k\}$, of which is defined by a RBF as
\begin{equation}
\begin{aligned}
    \label{eq:activation}
    \theta_{ij}(\bm{q})
    = \bm{e}^\frac{-||\bm{q}-\bm{\mu}_{ij}||^2}{\sigma_{ij}^2}
\end{aligned}
\end{equation}
where $\bm{\mu}_{ij} \in \mathbb{R}^{n}$ is a vector representing the center value of the $j$-th neuron, and $\sigma_{ij}$ is the corresponding width value, the basis of which can be determined according to the rough mapping from joint space to configuration space of continuum robot and detailed in Section IV.
In Equation (\ref{eq:Jnn}), $\bm{E}_J \in \mathbb{R}^{m \times n}$ denotes a matrix representing the functional approximation error of the NNs learning, which is upper bounded and could decrease with the increase in the number of neurons $k$ \cite{lewis2002neuro}.

Using NNs approximation to the combined Jacobian matrix in Equation (\ref{eq:Jnn}), the measurable shape flow $\dot{\bm{x}}$ using FBG sensing in Equation (\ref{eq:modeling}) can be further parameterized as
\begin{equation}
\begin{aligned}
    \label{eq:shape_flow}
	\dot{\bm{x}} = 
	\begin{bmatrix}
        \bm{W}_1 \bm{\theta}_1 (\bm{q})
        & \bm{W}_2 \bm{\theta}_2 (\bm{q}) & \cdots
        & \bm{W}_m \bm{\theta}_{m} (\bm{q})
    \end{bmatrix}^{\intercal}
	\dot{\bm{q}}
	+ \bm{\delta}
\end{aligned}
\end{equation}
where $\bm{\delta} \in \mathbb{R}^{m}$ is a combined perturbation term that describes the modeling disturbances $\bm{d}$ together with NN approximation error $\bm{E}_J$ given by
\begin{equation}
\begin{aligned}
    \label{comdis}
    \bm{\delta} = \bm{E}_J \dot{\bm{q}} + \bm{d}
\end{aligned}
\end{equation}
Considering the boundedness of $\dot{\bm{q}}$, $\ddot{\bm{q}}$, $\bm{d}$, $\dot{\bm{d}}$, $\bm{E}_J$, and its time derivative $\dot{\bm{E}}_J$ \cite{lewis2002neuro}, we have the combined perturbation $\bm{\delta}$ and its time derivative $\dot{\bm{\delta}} \in \mathbb{R}^{m}$ are bounded from Equation (\ref{comdis}) as
\begin{equation}
\begin{aligned}
    \label{eq:disbound}
    \left\| \bm{\delta} \right\| \le b_{\delta 1}, \quad
    \| \dot{\bm{\delta}} \| \le b_{\delta 2}
\end{aligned}
\end{equation}
where $b_{\delta 1}$ and $b_{\delta 2}$ are positive constants.

\begin{figure}[t]
  \centering
  \includegraphics[width=\linewidth]{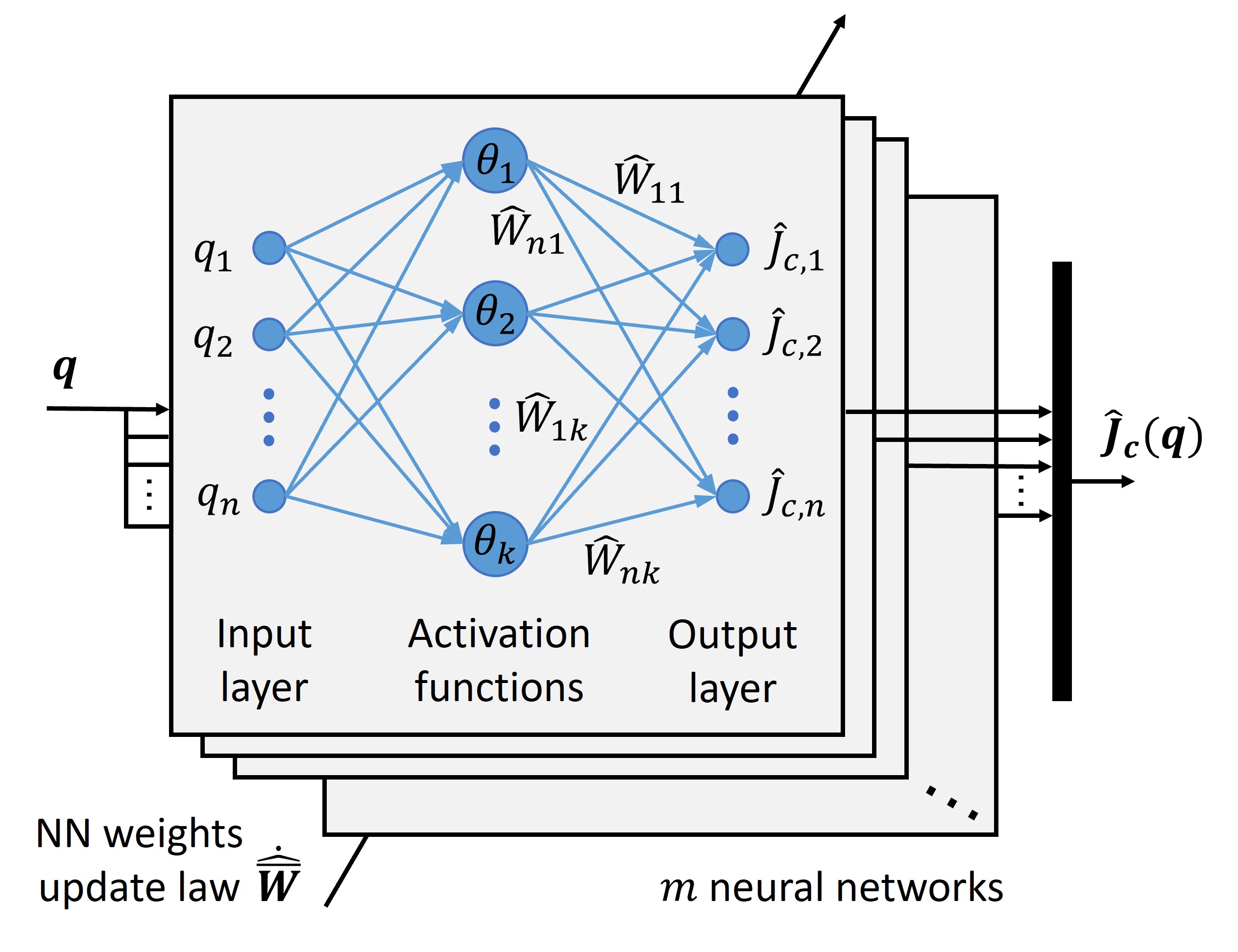}
  \caption{$m$ RBF NNs for Jacobian rows approximation, where for the $i$-th NN, $i \in \{1, 2, ..., m\}$: $q_{j_q}$, $j_q \in \{1, 2, ..., n\}$, represents the $j_q$-th element of $\bm{q}$, $\theta_{j_\theta}$, $j_\theta \in \{1, 2, ..., k\}$, is the $j_\theta$-th neuron of $\bm{\theta}_i$, $\widehat{J}_{c,j_J}$, $j_J \in \{1, 2, ..., n\}$, is the $j_J$-th element of the $i$-th row of $\widehat{\bm{J}}_{c}$, and $\widehat{W}_{j_J, j_\theta}$ is the $(j_J, j_\theta)$-th element of $\widehat{\bm{W}}_{i}$ updated by the adaptive law in Equation (\ref{eq:adaptive}).}
  \label{fig:nn}
\end{figure}

Since we do not have a priori knowledge of the deformation model of the continuum robot as well as the ideal NN weight matrices (i.e., $\bm{J}_c (\bm{q})$ and $\bm{W}_i, i \in \{1, 2, ..., m\}$, are all unknown), we estimate the combined Jacobian matrix denoted by $\widehat{\bm{J}}_c (\bm{q}) \in \mathbb{R}^{m \times n}$ using $m$ estimated NNs $\widehat{\bm{W}}_i \bm{\theta}_i (\bm{q}), i \in \{1, 2, ..., m\}$, as shown in Figure 2 and expressed by 
\begin{equation}
\begin{aligned}
    \label{eq:nnp2}
    \widehat{\bm{J}}_c (\bm{q}) =
    \begin{bmatrix}
        \widehat{\bm{W}}_1 \bm{\theta}_1 (\bm{q})
        & \widehat{\bm{W}}_2 \bm{\theta}_2 (\bm{q}) & \cdots
        & \widehat{\bm{W}}_{m} \bm{\theta}_{m} (\bm{q})
    \end{bmatrix}^{\intercal}
\end{aligned}
\end{equation}
where $\widehat{\bm{W}}_i \in \mathbb{R}^{n \times k}, i \in \{1, 2, ..., m\}$, is the estimated NN weight matrix.
To online tune the estimated NN weights and approximate the robot's model, a composite adaptation law for iterative evolution of the weights is proposed.
The prediction error of the shape flow is minimized by a robust shape flow predictor, and two vectors of NN weights for linear parametrization of the ideal and estimated combined Jacobian matrices are defined, which are all derived in detail as follows.

\subsection{Learning-Based Shape Flow Predictor}
To minimize the prediction error \cite{slotine1991applied} of shape flow for learning the unknown differential kinematics of continuum robot in Equation (\ref{eq:modeling}), we iteratively predict the shape flow denoted by $\dot{\widehat{\bm{x}}} \in \mathbb{R}^{m}$ and compare its difference with the measurable shape flow from FBG sensing results.
For that, a robust learning-based shape flow predictor composed of the NNs online approximation and sliding mode method \cite{dinh2010dynamic} is derived as
\begin{equation}
\begin{aligned}
    \label{eq:estimated_shape}
    \dot{\widehat{\bm{x}}} 
    & =
    \underbrace{
    \begin{bmatrix}
        \widehat{\bm{W}}_1 \bm{\theta}_1 (\bm{q})
        & \widehat{\bm{W}}_2 \bm{\theta}_2 (\bm{q}) & \cdots
        & \widehat{\bm{W}}_{m} \bm{\theta}_{m} (\bm{q})
    \end{bmatrix}^{\intercal}
    }_{\widehat{\bm{J}}_c (\bm{q})}
    \dot{\bm{q}}
    \\
    & \quad
    + \alpha_x \widetilde{\bm{x}}
    + \beta_x {\rm sat}(\widetilde{\bm{x}})
\end{aligned}
\end{equation}
where $\alpha_x$ and $\beta_x$ are positive constants; ${\rm sat} \left( \cdot \right)$ is a continuous saturation function as the smoothed implementation of sliding mode method, which can robustly remedy the system disturbances $\bm{d}$ and NN approximation error $\bm{E}_J$, while avoiding chattering phenomenon \cite{slotine1991applied}; and $\widetilde{\bm{x}} \in \mathbb{R}^{m}$ represents the shape estimation error defined by
\begin{equation}
\begin{aligned}
    \label{eq:estimate_error}
    \widetilde{\bm{x}} = \bm{x} - \widehat{\bm{x}}
\end{aligned}
\end{equation}
whose asymptotic convergence to zero can be guaranteed in the stability analysis.
Taking the time derivative of $\widetilde{\bm{x}}$ in Equation (\ref{eq:estimate_error}), the prediction error \cite{slotine1991applied} of shape flow defined as shape flow error $\dot{\widetilde{\bm{x}}} \in \mathbb{R}^{m}$, is derived from Equations (\ref{eq:estimated_shape}) and (\ref{eq:estimate_error}) as
\begin{equation}
\begin{aligned}
    \label{eq:x_estimator}
    \dot{\widetilde{\bm{x}}}
    & = 
    \dot{\bm{x}} - \dot{\widehat{\bm{x}}}
    \\
    & = 
    \dot{\bm{x}}
    - \widehat{\bm{J}}_c (\bm{q}) \dot{\bm{q}}
    - \alpha_x \widetilde{\bm{x}}
    - \beta_x {\rm sat}(\widetilde{\bm{x}})
\end{aligned}
\end{equation}
which is acquired by calculating the difference between the FBG measurement $\dot{\bm{x}}$ and the predictor $\dot{\widehat{\bm{x}}}$ using NN-estimated Jacobian $\widehat{\bm{J}}_c (\bm{q})$, actuation velocity $\dot{\bm{q}}$, together with the integral of the difference w.r.t. time, i.e., the shape estimation error $\widetilde{\bm{x}}$, whose initial value is set by $\widetilde{\bm{x}} = \bm{x}$ at $t = 0$ because we do not have a priori knowledge of it.

\subsection{Vectors of NN Weight Matrices}
We linearly parameterize the combined Jacobian in term of the NN weights to online update the unknown weights and learn the Jacobian for the controller.
Let us group the columns of $m$ ideal NN weight matrices $\bm{W}_i$ and $m$ estimated ones $\widehat{\bm{W}}_i$, $i \in \{1, 2, ..., m\}$, respectively, into two vectors of them as $\overline{\bm{W}} \in \mathbb{R}^{kmn}$ and $\widehat{\overline{\bm{W}}} \in \mathbb{R}^{kmn}$ given by
\begin{equation}
\begin{aligned}
    \label{eq:vector_W}
    \overline{\bm{W}} = 
    \begin{bmatrix}
        \bm{W}_{11}^{\intercal} & \bm{W}_{12}^{\intercal} & ...
        \bm{W}_{1k}^{\intercal} & \bm{W}_{21}^{\intercal} & ...
        & \bm{W}_{mk}^{\intercal}
    \end{bmatrix}^{\intercal}
\end{aligned}
\end{equation}
\begin{equation}
\begin{aligned}
    \widehat{\overline{\bm{W}}} = 
    \begin{bmatrix}
        \widehat{\bm{W}}_{11}^{\intercal} & \widehat{\bm{W}}_{12}^{\intercal} & ...
        \widehat{\bm{W}}_{1k}^{\intercal} & \widehat{\bm{W}}_{21}^{\intercal} & ...
        & \widehat{\bm{W}}_{mk}^{\intercal}
    \end{bmatrix}^{\intercal}
\end{aligned}
\end{equation}
where $\bm{W}_{ij}$ and $\widehat{\bm{W}}_{ij}$ are the $j$-th columns, $j \in \{1, 2, ..., k\}$,  of the $i$-th ideal and estimated NN weight matrix $\bm{W}_{i}$ and $\widehat{\bm{W}}_{i}$, $i \in \{1, 2, ..., m\}$,  respectively.

\subsection{Composite Adaptation for Online NN Weights Estimation}
Define two diagonal block matrices $\bm{\Theta}(\bm{q}) \in \mathbb{R}^{km \times m}$ and $\bm{Q} (\dot{\bm{q}}) \in \mathbb{R}^{kmn \times km}$, respectively, by grouping the activation functions $\bm{\theta}_i (\bm{q})$, $i \in \{1, 2, ..., m\}$, and $km$ actuator input $\dot{\bm{q}}$ as
\begin{equation}
\begin{aligned}
    &
    \label{eq:block_theta}
    \bm{\Theta}(\bm{q}) 
    = \mathrm{diag} ( \; 
    \underbrace{
    \bm{\theta}_{1} (\bm{q}), \ \bm{\theta}_{2} (\bm{q}), \ \cdots, \ \bm{\theta}_{m} (\bm{q})
    }_{\bm{\theta}_i (\bm{q}), \; i \in \{1, 2, ..., m\}} \; )
\end{aligned}
\end{equation}
\begin{equation}
\begin{aligned}
    &
    \label{eq:block_q}
    \bm{Q} (\dot{\bm{q}}) 
    = \mathrm{diag} ( \; 
    \underbrace{
    \dot{\bm{q}}, \ \dot{\bm{q}}, \ \cdots, \ \dot{\bm{q}}
    }_{km \ \dot{\bm{q}}} \; )
\end{aligned}
\end{equation}

The first term of the FBG-measured shape flow $\dot{\bm{x}}$ in Equation (\ref{eq:shape_flow}) approximated by the ideal NNs can be linearly parameterized using Equations (\ref{eq:vector_W}), (\ref{eq:block_theta}), and (\ref{eq:block_q}), given by
\begin{equation}
\begin{aligned}
    \label{eq:linear}
    & \begin{bmatrix}
        \bm{W}_1 \bm{\theta}_1 (\bm{q})
        & \bm{W}_2 \bm{\theta}_2 (\bm{q}) & \cdots
        & \bm{W}_m \bm{\theta}_{m} (\bm{q})
    \end{bmatrix}^{\intercal}
	\dot{\bm{q}}
	\\
	& \qquad = 
	\bm{\Theta}^{\intercal}(\bm{q})
	\begin{bmatrix}
        \bm{W}_1 
        & \bm{W}_2 & ...
        & \bm{W}_m 
    \end{bmatrix}^{\intercal}
	\dot{\bm{q}}
	\\
	& \qquad =
	\bm{\Theta}^{\intercal}(\bm{q}) 
	\bm{Q}^{\intercal}(\dot{\bm{q}}) 
	\overline{\bm{W}}
\end{aligned}
\end{equation}
as well as the NNs approximation of the predicted shape flow $\dot{\widehat{\bm{x}}}$ in Equation (\ref{eq:estimated_shape}) formulated as
\begin{equation}
\begin{aligned}
    \label{eq:estimated_linear}
    \begin{small}
    \setlength{\arraycolsep}{2pt}
    \begin{array}{cccc}
    \begin{bmatrix}
        \widehat{\bm{W}}_1 \bm{\theta}_1 (\bm{q})
        & \widehat{\bm{W}}_2 \bm{\theta}_2 (\bm{q}) & \cdots
        & \widehat{\bm{W}}_{m} \bm{\theta}_{m} (\bm{q})
    \end{bmatrix}^{\intercal}
    \end{array}
    \dot{\bm{q}}
    =
    \bm{\Theta}^{\intercal} (\bm{q})
	\bm{Q}^{\intercal} (\dot{\bm{q}}) 
	\widehat{\overline{\bm{W}}}
	\end{small}
\end{aligned}
\end{equation}
where $\bm{\Theta}^{\intercal}(\bm{q}) \bm{Q}^{\intercal}(\dot{\bm{q}})$ is also equivalent to $\bm{\Theta}^{\intercal}(\bm{q}) \otimes \dot{\bm{q}}^{\intercal}$ using Kronecker Product \cite{alambeigi2019scade}.
Then the measurable and predicted shape flow $\dot{\bm{x}}$ and $\dot{\widehat{\bm{x}}}$ can be expressed from the linearly parameterized formulations in Equations (\ref{eq:linear}) and (\ref{eq:estimated_linear}) as
\begin{equation}
\begin{aligned}
	\label{eq:x_linear}
	\dot{\bm{x}} 
	= \bm{\Theta}^{\intercal}(\bm{q}) 
	\bm{Q}^{\intercal}(\dot{\bm{q}}) 
	\overline{\bm{W}}
	+ \bm{\delta}
\end{aligned}
\end{equation}
\begin{equation}
\begin{aligned}
	\label{eq:estimatedx_linear}
	\dot{\widehat{\bm{x}}} 
	= \bm{\Theta}^{\intercal}(\bm{q}) 
	\bm{Q}^{\intercal}(\dot{\bm{q}}) 
	\widehat{\overline{\bm{W}}}
	+ \alpha_x \widetilde{\bm{x}}
    + \beta_x {\rm sat}(\widetilde{\bm{x}})
\end{aligned}
\end{equation}
and the difference between the ideal and estimated NNs approximations to the combined Jacobian matrices multiplying the actuation velocity $\dot{\bm{q}} $, i.e., the difference between $\bm{J}_c (\bm{q}) \dot{\bm{q}}$ and $\widehat{\bm{J}}_c (\bm{q}) \dot{\bm{q}}$, is calculated by
\begin{equation}
\begin{aligned}
    \label{eq:JJW}
    &
    \bm{J}_c (\bm{q}) \dot{\bm{q}} 
    - \widehat{\bm{J}}_c (\bm{q}) \dot{\bm{q}}
    = \bm{\Theta}^{\intercal}(\bm{q}) \bm{Q}^{\intercal}(\dot{\bm{q}})
    \widetilde{\overline{\bm{W}}}
    + \bm{E}_J \dot{\bm{q}}
\end{aligned}
\end{equation}
where $\widetilde{\overline{\bm{W}}} = \overline{\bm{W}} - \widehat{\overline{\bm{W}}} \in \mathbb{R}^{kmn}$ denotes the estimation error of NN weights in vector form.
From Equations (\ref{eq:x_linear}) and (\ref{eq:estimatedx_linear}), we can further derive the linearly parameterized shape flow error $\dot{\widetilde{\bm{x}}}$ as
\begin{equation}
\begin{aligned}
    \label{eq:estimator}
    \dot{\widetilde{\bm{x}}}
    =
    \bm{\Theta}^{\intercal}(\bm{q}) \bm{Q}^{\intercal}(\dot{\bm{q}})
    \widetilde{\overline{\bm{W}}}
    + \bm{\delta}
    - \alpha_x \widetilde{\bm{x}}
    - \beta_x {\rm sat}(\widetilde{\bm{x}})
\end{aligned}
\end{equation}

In addition to the above derivations for the adaptive update of the NN weights, we introduce the filtered shape estimation error $\bm{r}_x \in \mathbb{R}^{m}$ from Equation (\ref{eq:estimator}) defined by
\begin{equation}
\begin{aligned}
    \label{eq:rxW}
    \bm{r}_x 
    = \dot{\widetilde{\bm{x}}} + \alpha_x \widetilde{\bm{x}}
    = \bm{\Theta}^{\intercal}(\bm{q}) \bm{Q}^{\intercal}(\dot{\bm{q}})
    \widetilde{\overline{\bm{W}}}
    + \bm{\delta}
    - \beta_x {\rm sat}(\widetilde{\bm{x}})
\end{aligned}
\end{equation}
which can be minimized using the following adaptation law and is available to ensure 
the asymptotic regulation of the shape flow error $\dot{\widetilde{\bm{x}}}$ compared with other forms difficult for this objective, e.g., simply $\dot{\widetilde{\bm{x}}}$.

Consequently, the vector of the estimated NN weights $\widehat{\overline{\bm{W}}}$ can be iteratively updated by using a composite adaptation rule modified from Slotine–Li method in \cite{slotine1991applied} and proposed as
\begin{equation}
\begin{aligned}
    \label{eq:adaptive}
    \dot{\widehat{\overline{\bm{W}}}}
	=
	& \bm{\Gamma}^{-1}_W {\rm proj}
	\Big \{
	k_e \bm{Q} (\dot{\bm{q}}) \bm{\Theta} (\bm{q}) 
	\bm{e}
	\\
	& +
	k_x \bm{Q} (\dot{\bm{q}}) \bm{\Theta} (\bm{q}) 
	\widetilde{\bm{x}}
	+
	k_r \bm{Q} (\dot{\bm{q}}) \bm{\Theta} (\bm{q}) 
	\bm{r}_x
    \Big \}
\end{aligned}
\end{equation}
where $k_e$, $k_x$, and $k_r$ are positive constants, 
$\bm{\Gamma}_W$ is a positive definite and diagonal gain matrix, 
$\bm{e} = \bm{x} - \bm{x}_d \in \mathbb{R}^{m}$ denotes the FBG-shape feedback error (control error),
$\bm{x}_d \in \mathbb{R}^{m}$ represents the desired shape of the continuum robot, and $\rm{proj} \{ \cdot \}$ is a projection operator to guarantee the estimated NN weights bounded \cite{krstic1995nonlinear}.
In Equation (\ref{eq:adaptive}), the first term of the shape control error is for the compensation of the difference between the estimated and real Jacobian matrices.
The second term of the shape estimation error is to account for a cross term in the stability analysis.
The third term about the filtered shape estimation error is for the online gradient descending minimization of the errors about the estimated model parameters.

\begin{figure}[t]
  \centering
  \includegraphics[width=\linewidth]{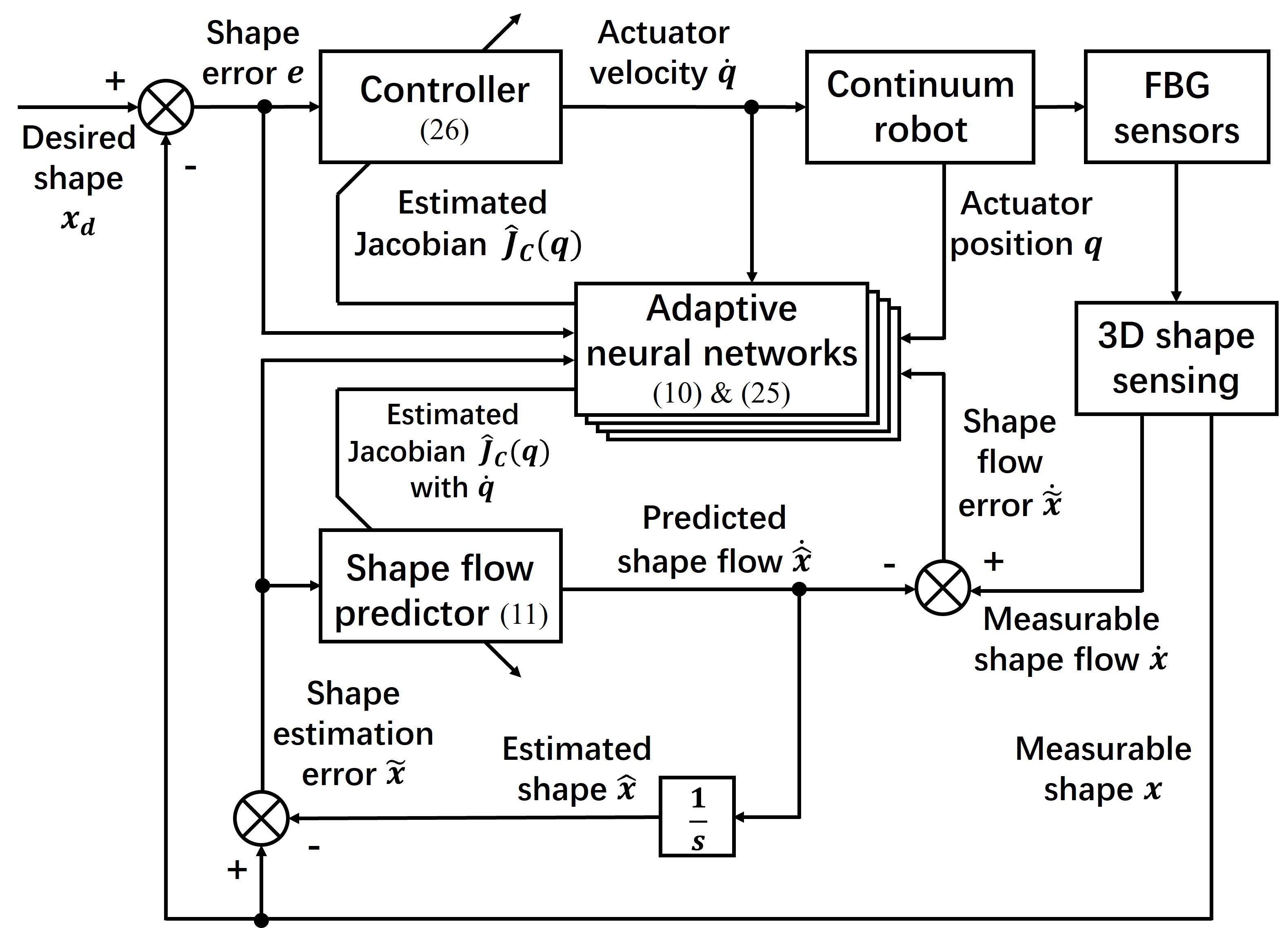}
  \caption{Block diagram of the closed-loop system under the proposed control algorithm.}
  \label{fig:controller}
\end{figure}

\subsection{Controller Design and Stability Analysis}
Constructing the estimated combined Jacobian $\widehat{\bm{J}}_c (\bm{q})$ from the NNs online learning by Equation (\ref{eq:nnp2}), the controller is designed as
\begin{equation}
\begin{aligned}
    \label{eq:control}
    \dot{\bm{q}} =
    - k_c \widehat{\bm{J}}^{+}_c (\bm{q}) \bm{e} 
    - k_s \widehat{\bm{J}}^{+}_c (\bm{q}) {\rm sat}(\bm{e})
\end{aligned}
\end{equation}
where $k_c$ and $k_s$ are positive constants, $\widehat{\bm{J}}^{+}_c (\bm{q}) \in \mathbb{R}^{n \times m}$ represents the Moore–Penrose pseudo-inverse of $\widehat{\bm{J}}_c (\bm{q})$, and $\bm{e}$ is the shape control error.
In Equation (\ref{eq:control}), the second term using a continuous saturation function is designed to eliminate the perturbations $\bm{\delta}$ in the measurement flow.
The implementation of the proposed algorithm is detailed in Appendix A, and the block diagram of the close-loop system under our controller with online learning is demonstrated in Figure 3.

\begin{figure}[!ht]
  \centering
  \includegraphics[width=0.95\linewidth]{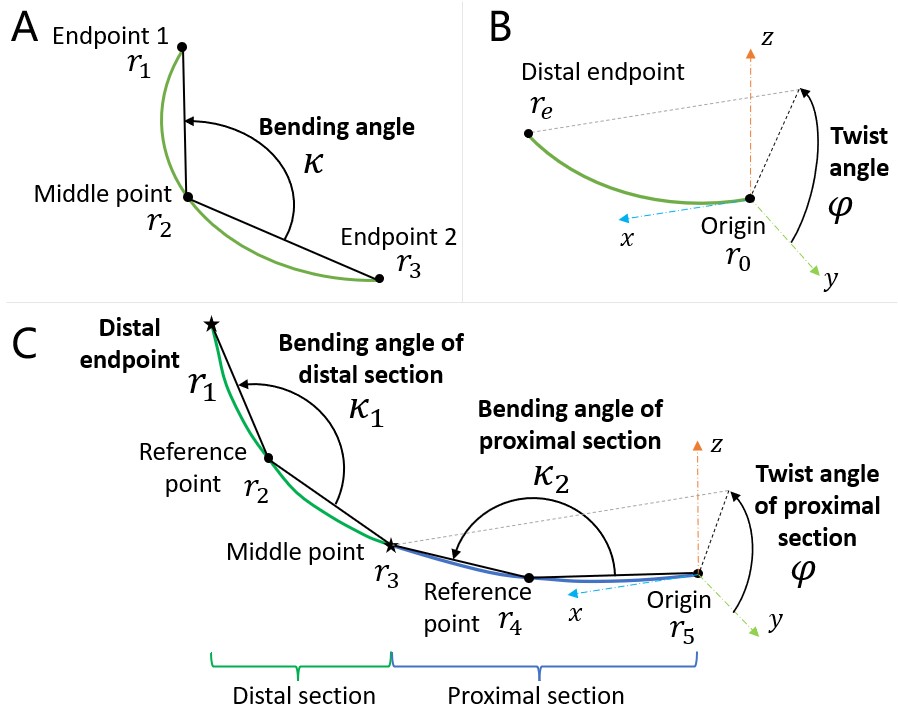}
  \caption{3-D shape features: (A) bending angle, (B) twist angle, and (C) distal endpoint position with bending/twist angles (DEP-BTA) for multi-section manipulator.}
  \label{fig:shape}
\end{figure}

Under the proposed shape flow predictor, composite adaptive algorithm, and velocity controller, the stability of the closed-loop system is guaranteed and proved in Appendix B, which gives rise to all the convergences of the shape control error $\bm{e}$, shape estimation error $\widetilde{\bm{x}}$, and shape flow error $\dot{\widetilde{\bm{x}}}$ to zeros as $t \rightarrow \infty$, i.e., 
\begin{equation}
\begin{aligned}
    &
    \lim\limits_{t \to \infty} \bm{e} = 0, 
    \quad
    \lim\limits_{t \to \infty} \widetilde{\bm{x}} = 0,
    \quad
    \lim\limits_{t \to \infty} \dot{\widetilde{\bm{x}}} = 0
\end{aligned}
\end{equation}
together with the convergence of estimated NN weights $\widehat{\bm{W}}_i$, $i \in \{1, 2, ..., m\}$, and guaranteed NNs learning of the unknown model, if the predictor parameters $\alpha_x$, $\beta_x$, and controller parameter $k_s$ are chosen such that
\begin{equation}
\begin{aligned}
    \label{eq:sufficient}
    k_s \ge b_{\delta 1}, \quad
    \beta_x \ge b_{\delta 1} + \frac{b_{\delta 2}}{\alpha_x}
\end{aligned}
\end{equation}
where $b_{\delta 1}$ and $b_{\delta 2}$ are positive constants in Equation (\ref{eq:disbound}).

\section{Experimental Setup}
To evaluate the proposed control algorithm, a comprehensive experimental study has been performed.
We introduce the setup in this section, including general shape features, two platforms, FBG sensing system, and initialization.

\subsection{General Shape Features}
Two 3-D shape features including point's position and bending/twist angles (called BTA) are defined, which can be composed to globally describe the shape of continuum robot.
Point's position refers directly to the Cartesian position $\bm{r}_i = \begin{bmatrix} x_i & y_i & z_i \end{bmatrix}^{\intercal} \in \mathbb{R}^{3}, i \in \{1, 2, ..., l\}$, selected on the continuum robot, which can also be utilized to construct the following features.
BTA consists of bending angle $\kappa$ and twist angle $\varphi$, which describe the deflection and torsion of one robot's section, respectively.
Bending angle $\kappa \in \mathbb{R}$, for $0^{\circ} < \kappa \le 180^{\circ}$, is a scalar that describes the angle between two intersecting lines constructed from three reference points $\bm{r}_i, i \in \{1, 2, 3\}$, as shown in Figure 4A by
\begin{equation}
\begin{aligned}
    \label{eq:bending}
    \begin{small}
    \kappa = {\rm arccos} \left(
    \frac{(\bm{r}_1 - \bm{r}_2) \cdot (\bm{r}_3 - \bm{r}_2)}
    {\left\| \bm{r}_1 - \bm{r}_2 \right\| \left\| \bm{r}_3 - \bm{r}_2 \right\|}
    \right)
    \end{small}
\end{aligned}
\end{equation}
Twist angle $\varphi \in \mathbb{R}$, for $-180^{\circ} < \varphi \le 180^{\circ}$, describes the rotation about the x-axis of the base frame as shown in Figure 4B given by
\begin{equation}
\begin{aligned}
    \label{eq:twist}
    \varphi = {\rm atan2} \left(z_e, y_e\right)
\end{aligned}
\end{equation}
where $\bm{r}_e = \begin{bmatrix} x_e & y_e & z_e \end{bmatrix}^{\intercal} \in \mathbb{R}^{3}$ is the section's endpoint.

\subsection{Robotic-Assisted Colonoscope System}
We integrated a FBG-equipped robotic-assisted colonoscope system (called RACS here) previously reported in \cite{lu2021robust}, of which the distal continuum mechanism with 120 mm working length 
was employed to perform the 
experiments, as shown in Figure 5A.
A dial drive module was designed to control the deflection motions of the continuum mechanism with 2 DOFs (i.e., $n = 2$).
To globally describe the shape of the distal continuum mechanism, two features are defined including (a) two points' positions and (b) BTA.
The feature of two points' positions $\bm{x} = \begin{bmatrix} \bm{r}_{e1}^{\intercal} & \bm{r}_{e2}^{\intercal} \end{bmatrix}^{\intercal} \in \mathbb{R}^{6}$ is constructed by stacking the Cartesian position of the distal endpoint $\bm{r}_{e1} \in \mathbb{R}^{3}$ and that of the middle point $\bm{r}_{e2} \in \mathbb{R}^{3}$, 
and the feature BTA $\bm{x} = \begin{bmatrix} \kappa & \varphi \end{bmatrix}^{\intercal} \in \mathbb{R}^{2}$ consists of the bending and twist angles.

\begin{figure*}[!ht]
  \centering
  \includegraphics[width=0.9\linewidth]{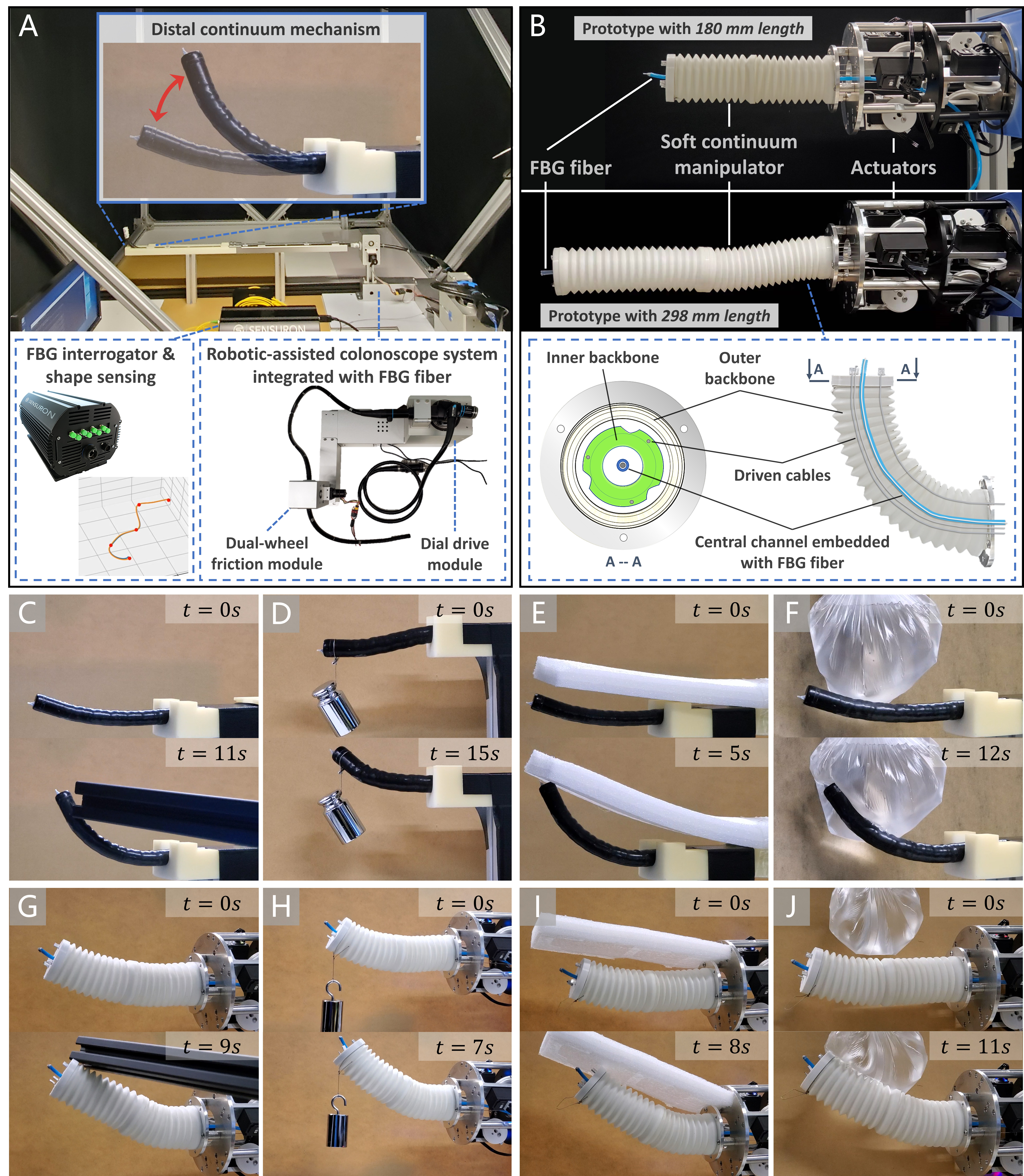}
  \caption{Experimental platforms integrated with FBG fibers: (A) robotic-assisted colonoscope system (RACS); (B) multi-section extensible soft continuum manipulators (SCM). Task setups using two platforms in unstructured environments with (C)/(G) a rigid rod, (D) 100 g/(H) 200 g payload, (E)/(I) elastic foam, and (F)/(J) a soft water bag.}
  \label{fig:setup}
\end{figure*}

\subsection{Multi-Section Extensible Soft Continuum Manipulator}
A soft continuum manipulator (called SCM here) with a 180 mm working length, two extensible sections, and 6 DOFs in total (i.e., $n = 6$), has been integrated and deployed for the experiments as shown in Figure 5B.
Its extensible continuum section, including outer and inner backbones fabricated by 
soft materials (CR-TPU) 
and capable of omni-directional deflections under variable stiffness/length, was previously deployed as a wrist component of a 
robotic platform for oropharyngeal swabs sampling \cite{chen2021tele}.
The deflection motions of each soft section are driven by three cables and actuated by servo motors.
Filled with small particles in the inner chamber (backbone) of the soft section, its stiffness/length can be adjusted by pulling/pushing three cables together \cite{zhou2020adaptive}.

Besides two points' positions $\bm{x} \in \mathbb{R}^{6}$ as a type of shape feature, another shape description called distal endpoint position with bending/twist angles (called DEP-BTA) $\bm{x} \in \mathbb{R}^{6}$ is defined by stacking the Cartesian position of distal endpoint $\bm{r}_1 \in \mathbb{R}^{3}$ and three angles $\bm{\xi} \in \mathbb{R}^{3}$, including two bending angles of distal section $\kappa_1$ and proximal section $\kappa_2$, and one twist angle of proximal section $\varphi$, as shown in Figure 4C by
\begin{equation}
\begin{aligned}
    \label{eq:soft_shape}
    \bm{x}
    = \begin{bmatrix} \bm{r}_1^{\intercal} & \bm{\xi}^{\intercal} \end{bmatrix}^{\intercal}, 
    \quad
    \bm{\xi} = \begin{bmatrix} \kappa_1 & \kappa_2 & \varphi \end{bmatrix}^{\intercal}
\end{aligned}
\end{equation}
where $\bm{r}_1$ can locate the desired distal endpoint and determine the total length of soft robot, $\kappa_1$ and $\kappa_2$ are utilized to describe the deflections of both sections, and $\varphi$ can determine the direction of the intersection point of two sections.

\subsection{FBG Shape Sensing System}
To acquire the 3-D shape of the continuum robots, two multi-core all-grating FBG fibers with outer diameters of 190 $\mathrm{\mu m}$ were separately integrated into the inner channels of two robots.
The fiber has 1 center core and 3 outer cores positioned at $120^{\circ}$ with 37 $\mathrm{\mu m}$ distance to the center.
We used an interrogator (RTS125+, Sensuron) to simultaneously decode strains from 45 and 55 FBG-sets of two robots, respectively, with both 30 Hz acquisition frequency and 3.3 mm spatial resolution.
Robust and accurate shape sensing results can be obtained by employing a modified version of the previous algorithm \cite{lu2021robust} with 25 Hz sensing frequency.
The frequency of the overall closed-loop control system 
is 20 Hz, which is 
comparable with those in the related works \cite{alambeigi2019scade, alambeigi2020versatile}.

\subsection{Initialization}
The parameters adopted in the study were tuned based on the preliminary experiments providing the desired performance with convergences of shape control and estimation errors to zeros.
In the experiments using RACS, the parameters were set as follows: $\alpha_x = 0.3$, $\beta_x = 0.04$, $k_e = 0.01$, $k_x = 0.01$, $k_r = 0.2$, $\bm{\Gamma}_W^{-1} = 0.1 \times \bm{I}_{18m \times 18m}$, $k_c = 0.32$, and $k_s = 0.04$.
For online learning, we used 9 neurons for each NN (i.e., $k = 9$) and RBFs as their activation functions.
The centers $\bm{\mu}_{ij} \in \mathbb{R}^{2}$ of the $i$-th NN, $i \in \{1, 2\}$, were initialized by randomly sorting the columns of the chosen basis $\bm{\mu}_{0j} \in \mathbb{R}^{2}$, $j \in \{1, 2, ..., 9\}$, and so were the corresponding widths $\sigma_{ij}$ from $\sigma_{0j} $. 
which aims to prevent the rows of the combined Jacobian from being linearly dependent.
This can consequently avoid the Jacobian ill-conditioned or even singular, eliminating overspeed motions. 
For SCM, the parameters were 
set as:
$\alpha_x = 0.6$, $\beta_x = 0.12$, 
$k_e = 0.01$, $k_x = 0.01$, $k_r = 0.2$, $\bm{\Gamma}_W^{-1} = \bm{I}_{78m \times 78m}$, 
$k_c = 0.02$, and $k_s = 0.01$, 
13 neurons were utilized for each NN (i.e., $k = 13$), and
the center and width values 
were initialized in the same way as they were when using RACS.

Given that there was no requirement of a priori identification for the robot's model, we initialized the estimated shape by zeros, i.e., $\widehat{\bm{x}} = \bm{0}_{m \times 1}$.
To enable fast learning convergence and improve the transient performance of the proposed 
algorithm, a normalized initialization method \cite{he2015delving} was utilized to initialize the NN weights.
Then, we conducted a few iterations of continuous slow motions in a small region around the initial configuration in the free environments \cite{navarro2016automatic}, which can speed up the convergence during 3-D shape servoing.

\begin{figure}[!ht]
  \centering
  \includegraphics[width=0.95\linewidth]{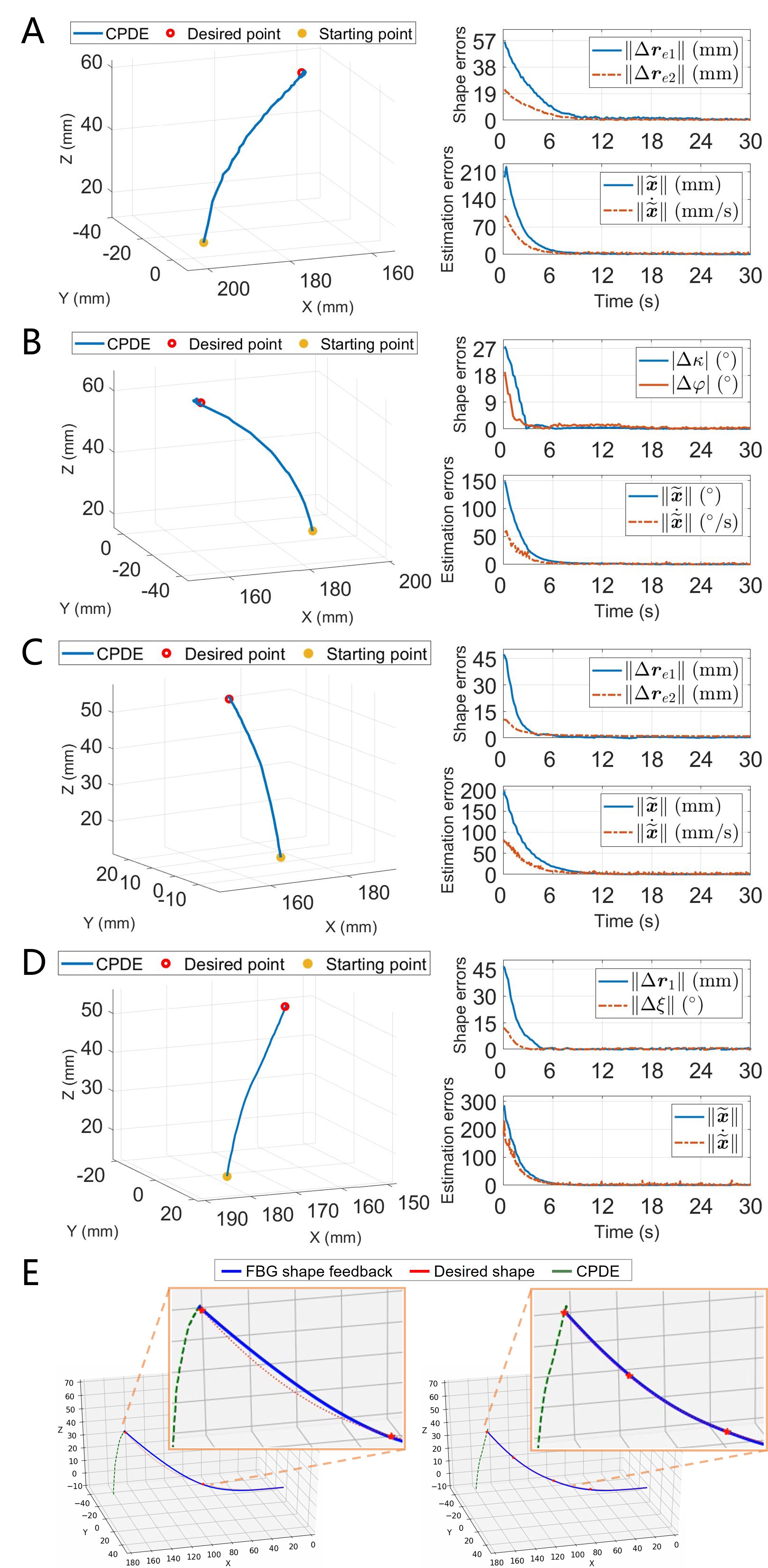}
  \caption{CPDE (left) and shape errors (right) of experiments using RACS in free environments, with (A) two points and (B) BTA as shape features. Those using SCM with (C) two points and (D) DEP-BTA as shape features, and (E) final shapes described by two points (left) and DEP-BTA (right).}
  \label{fig:expt1}
\end{figure}

\begin{figure*}[!ht]
  \centering
  \includegraphics[width=0.97\linewidth]{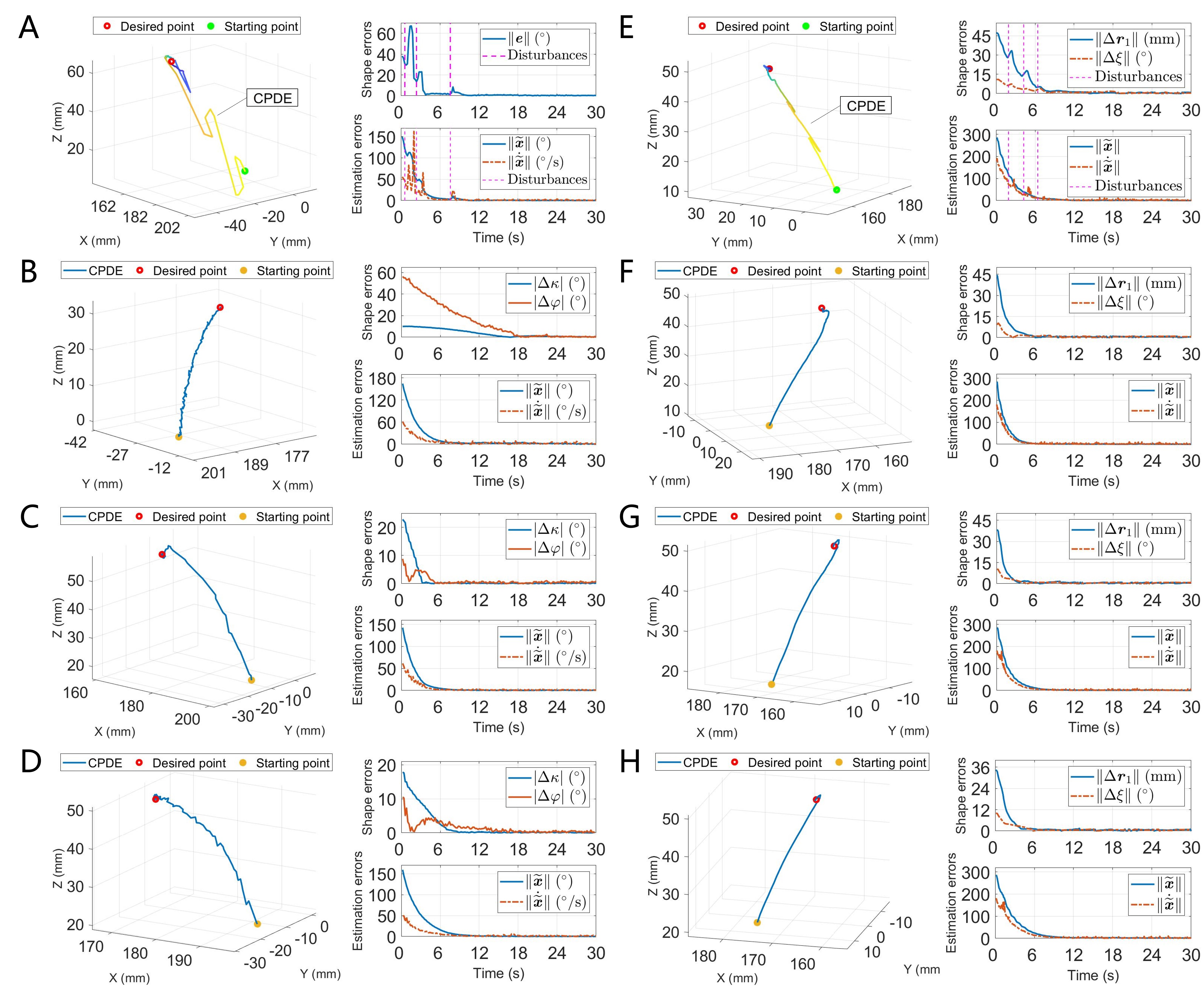}
  \caption{CPDE (left) and shape errors (right) of experiments in unstructured environments, including the tests using RACS with (A) a rigid rod, (B) 100 g payload, (C) elastic foam, and (D) a soft water bag, and those using SCM with (E) a rigid rod, (F) 200 g payload, (G) elastic foam, and (H) a soft water bag.}
  \label{fig:expt2}
\end{figure*}

\section{Experimental Results}
Experimental validations (29 experiments in total) were performed to evaluate the performance of the proposed algorithm, including shape servoing tasks 
using two continuum robots in free and unstructured environments, validation for effectiveness of NNs online learning, validation on a longer continuum robot, validation with unreachable desired shape, comparison experiments 
, and case studies in medical contexts.
The results including Cartesian path of the robot's distal endpoint (CPDE), as well as 2-norms of shape errors, i.e., those of shape control errors $\|\bm{e}\|$, shape estimation errors $\|\widetilde{\bm{x}}\|$, and shape flow errors $\|\dot{\widetilde{\bm{x}}}\|$, are recorded and presented.
In all figures showing CPDE and shape errors, the left plots show the resulting CPDE.
Except special explanations, the top-right images demonstrate the shape control errors, where 
Cartesian position errors $\Delta \bm{r}_{e1}, \Delta \bm{r}_{e2} \in \mathbb{R}^{3}$ of the endpoint $\bm{r}_{e1}$ and middle point $\bm{r}_{e2}$ are presented for the errors described by two points.
For the control errors described by BTA (using RACS), 
bending and twist errors $\Delta \kappa, \; \Delta \varphi \in \mathbb{R}$ are presented, and position error of distal endpoint $\Delta \bm{r}_{1} \in \mathbb{R}^{3}$ with bending/twist error $\Delta \bm{\xi} \in \mathbb{R}^{3}$ are shown for those described by DEP-BTA (using SCM).
The bottom-right plots show the shape estimation error $\widetilde{\bm{x}}$ and shape flow error $\dot{\widetilde{\bm{x}}}$.
We also attach Supplementary Video S1 emphasizing the exterior views of the continuum robots and simultaneous FBG-shape convergences.


\subsection{Shape Servoing Tasks in Free Environments}
We first conducted the shape control tasks in free space using RACS and SCM, respectively, and the resulting CPDE and shape errors are depicted in Figure 6, where two features were used to describe the shape of each robot.
In the left plots of Figure 6A and 6B 
using RACS, distal endpoints can approach the desired positions with both shape features, 
though endpoint positioning was not processed when described by BTA.
The right images indicate the asymptotic minimization of shape control and estimated errors regardless of different shape descriptions, showing that the 3-D shape of continuum mechanism can automatically converge to the desired one.

Given that the length of SCM can be adjusted, 
the arc lengths of its initial and desired shapes are different in all experiments, where the desired endpoint position is outside the initial workspace.
The asymptotic convergences of shape errors in Figure 6C and 6D show that the robot's shape described by two features can be controlled to the desired configurations.
Notice that the final shape described by two points is different from the desired one as shown in 
Figure 6E, where the distal section is nearly stretched to a straight line, while that described by DEP-BTA incorporating endpoint positioning and 
deflections imposes a desired configuration.

\subsection{Shape Servoing Tasks in Unstructured Environments}
These experiments were performed using two continuum robots in unstructured environments 
as shown in Figure 5C-J, including the tasks with a rigid rod, 100 g and 200 g payloads, foam, and a soft water bag.
The results are presented in Figure 7, where two features
were utilized to describe the robot shapes, 
and Supplementary Video S1 provides those using two points.
Note that the initial and desired lengths of SCM are different in all experiments, during which the robot has to be extended to impose the desired shape.

Figure 7A and 7E indicate the first experimental results in collisions with the rigid rod, during which we performed three random collisions to the distal parts of two continuum robots 
using the rod as unexpected 
disturbances.
Notice that CPDE with obvious perturbations in the left images are plotted by gradient colors.
We can find the sudden increases of the shape errors in the right figure,
where the pink lines represent the time instants encountering the collisions.
It is worth observing that the errors can keep converging after each collision,
which is contributed by the robustness of shape flow predictor and controller, as well as the adaptability of NNs online learning despite intense 
perturbations.
We then evaluated the algorithm by applying two masses of 100 g and 200 g weights, respectively, which can be treated as extra unknown payloads on the tips of two continuum robots as shown in Figure 5D and 5H.
The initial shapes suffer unexpected and obvious deformation under unknown payloads as compared with 
others in Figure 5.
The results in Figure 7B and 7F illustrate that CPDE can approach the desired point, and all the shape control and estimation errors are asymptotically minimized to zeros in the presence of unknown payload.

The third experiments were performed with collisions 
with an elastic foam as shown in Figure 5E and 5I, 
which mimic interacting with elastic tissues during surgery.
The robot model after the collision is different from that initially updated in free space, which becomes inaccurate during the interaction.
The left plots of Figure 7C and 7G depict that CPDE outputs a smooth path in free space, and it approaches the desired configuration by 
online approximation to the interaction model after the collision.
Such performance is also 
illustrated in the right images, where the shape errors asymptotically converge to zeros.
In the fourth experiments, a soft water bag of 340 g weight hangs over the continuum robots as shown in Figure 5F and 5J.
The robot has to lift the bag regarded as a soft tissue and move to the desired configuration, and it is concerned difficult to obtain a priori accurate model of the robot in this scenario considering multiple factors that can affect the robot's behavior (e.g., payloads and frictions).
The resulting CPDE and asymptotic convergences of shape errors to zeros are shown in Figure 7D and 7H.
Specifically, the shape errors decrease rapidly in free space, and then converge to zeros during the interaction, which is contributed by the fast adaptation to unknown disturbances under the proposed controller.
These results demonstrate that our algorithm can online learn the time-varying behavior of continuum robot, adaptively conform to unstructured environments, and simultaneously control the robot approaching the desired shape.

\begin{figure}[!ht]
  \centering
  \includegraphics[width=0.97\linewidth]{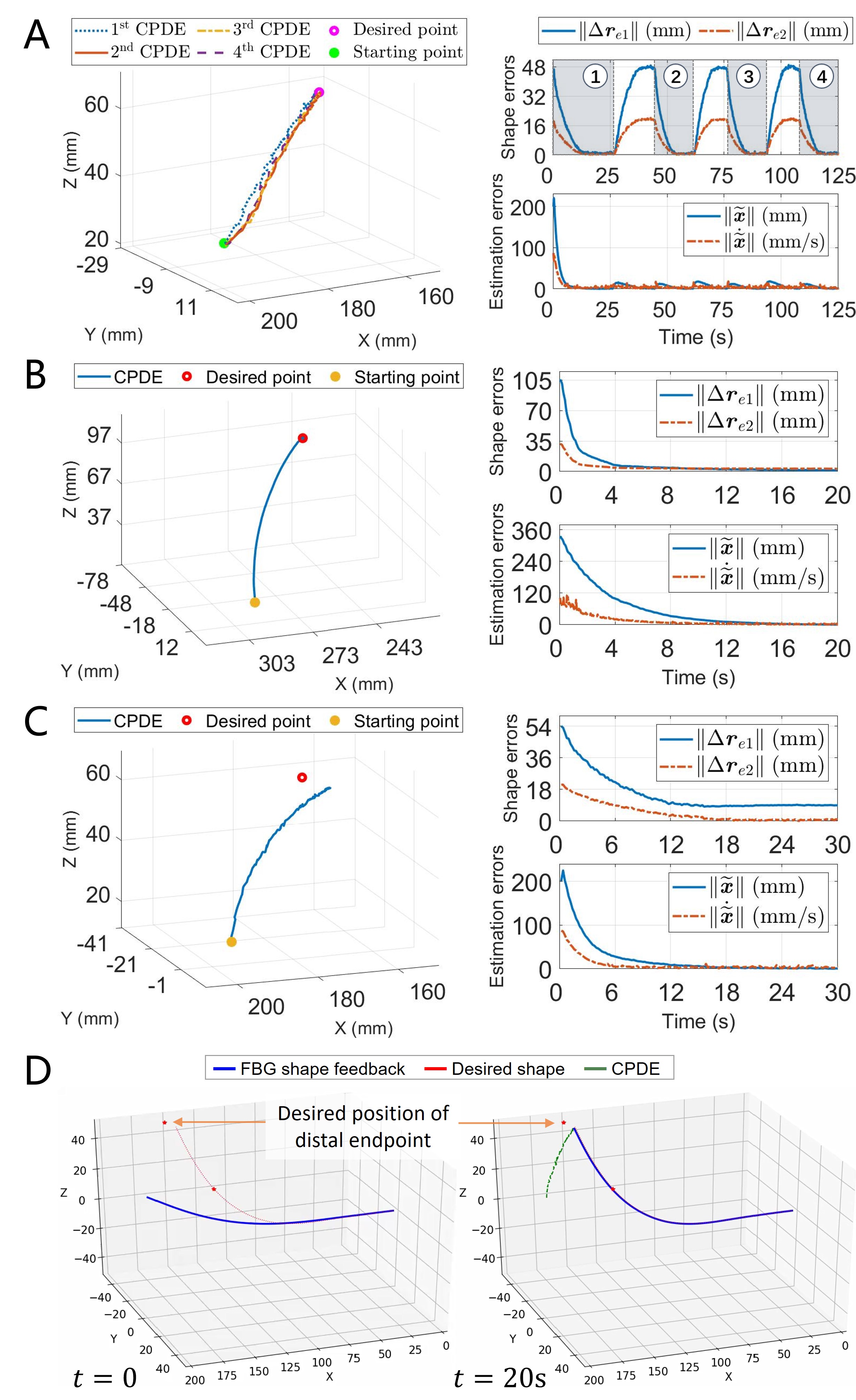}
  \caption{CPDE (left) and shape errors (right) of validations: (A) effectiveness of NNs online learning. (B) task on longer soft continuum manipulator. (C) task with unreachable desired shape and (D) its initial (left) and final (right) shapes.}
  \label{fig:expt3}
\end{figure}

\subsection{Effectiveness of NNs Online Learning}
To prove that the proposed algorithm is able to learn the 
combined Jacobian $\bm{J}_c (\bm{q})$, five shape servoing tasks using RACS 
were performed, and each task was repeated four times.
After the first completion of each task, the afterward experiments were repeated and started by utilizing the weight matrices 
updated from the previous completion.
Note that during each task, the 
shape was controlled back to the same initial configuration after each completion, the processes of which are indicated by the white gaps in the top-right plot of Figure 8A, while those of the repeated experiments are marked and presented in the shade.
From the average convergence time (in terms of repeated times) of five tasks in Table \ref{table:task3} as well as the asymptotic convergences of shape errors in 
Figure 8A, 
it can be noticed that the latter three convergences are faster than the first one (by around 28.25\%) 
during each task and they take a similar amount of time.
Meanwhile, the distal endpoint has similar paths during the latter few convergences as shown in the left image.
These results indicate that the proposed shape flow predictor together with the composite adaptation law have updated the weight matrices to steady values after the first attempt, and the learned Jacobian during the afterward repeated experiments only depends on the robot configuration.

\begin{table}[!hb]
\caption{\centering \textbf{Average convergence time (s)} of experiments to validate the effectiveness of NNs online learning}
\label{table:task3}
\centering
\small
\begin{tabular}{c|cccc}
\hline
Repeat times & 1     & 2     & 3     & 4     \\ \hline
Average time (s)     & 15.33 & 11.00 & 11.40 & 10.87 \\ \hline
\end{tabular}
\end{table}

\subsection{Validation on Longer Continuum Robot}
This experiment was performed using a soft continuum manipulator newly designed with two longer sections, the length of which was increased 
to 298 mm as shown in Figure 5B, and 90 FBG-sets 
were accordingly utilized for 3-D shape 
feedback.
Due to the variable length of soft manipulator, its initial and desired lengths are different.
Figure 8B demonstrates the resulting CPDE and asymptotic convergences of shape errors described by two points, showing that the proposed algorithm is generic and feasible for longer continuum robot.
The validation on another longer robot can refer to the afterward phantom experiment using RACS with 400 mm length.

\subsection{Validation with Unreachable Desired Shape}
An experiment using RACS was conducted to validate the proposed method with an unreachable desired shape, where the desired position of distal endpoint is outside its workspace as shown in Figure 8D.
CPDE and shape errors are depicted in Figure 8C,
from which we can observe a steady-state error in the 
endpoint position error $\| \Delta \bm{r}_{e1} \|$
since its desired position is not feasible.
Although the distal endpoint cannot be controlled to the desired position, the right plot of Figure 8D demonstrates that the shape of continuum mechanism can still approach the desired one by using the proposed algorithm.

\subsection{Comparison Experiments}
We conducted four tasks to compare the performance of the proposed 
approach with two model-based methods using a fixed model and a constant curvature model, respectively, in terms of CPDE, shape control errors $\|\bm{e}\|$, and estimation errors $\|\widetilde{\bm{x}}\|$.
By deploying RACS with BTA as shape feature, the first two experiments were compared with a controller using a constant combined Jacobian $\widehat{\bm{J}}_c (\bm{q})$ estimated around the starting point in free space, the purpose of which is to compare with a method that only utilizes a fixed model \cite{navarro2016automatic}.
Their results 
can be found in Supplementary Video S1, where the fixed-model method demonstrates oscillation even divergence, while ours shows asymptotic convergences and robustness against unknown conditions.

\begin{figure}[!ht]
  \centering
  \includegraphics[width=0.95\linewidth]{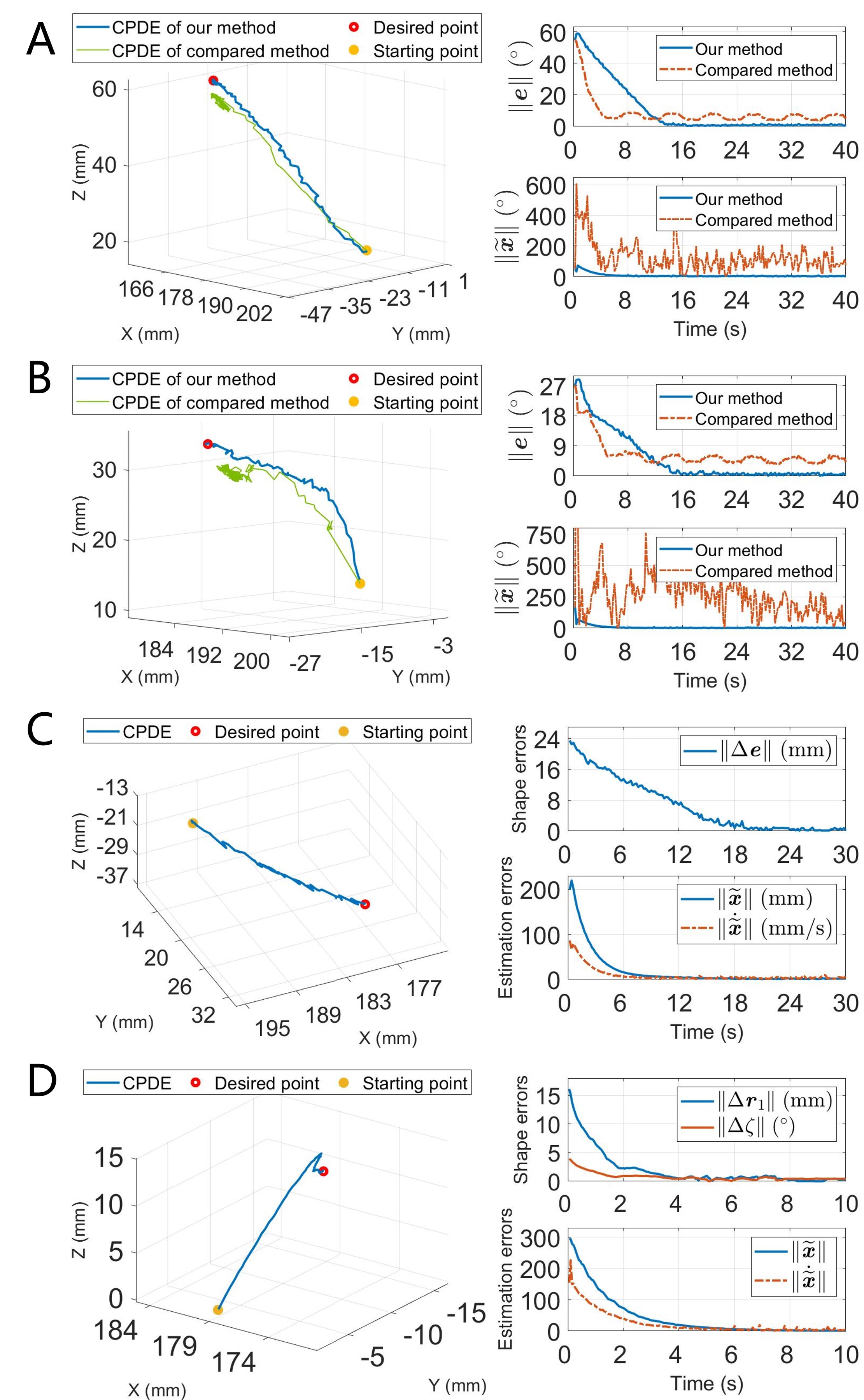}
  \caption{Comparison experiment results with model-based method using constant curvature: CPDE (left), and shape control and estimation errors (right) with (A) internal uncertainties and (B) external disturbances in addition. Results of phantom experiments: CPDE (left) and shape errors (right) of (C) shape control for robotic colonoscopy and (D) OP swab sampling.}
  \label{fig:expt4}
\end{figure}

The next two experiments were performed by comparing with a controller using a constant curvature model \cite{webster2010design}, where the robot shape was described by two points.
The first test was processed by injecting the same unknown uncertainties into the actuator command input.
Figure 9A demonstrate that although the results using the model-based method without modeling the uncertainties converges faster around the starting point than ours, its shape is not controlled to the desired one but oscillates in a region after 5.6 s.
Our algorithm initially costs a few more iterations for adaption to the time-varying model with unknown uncertainties, while its errors show asymptotic minimization to zeros at 11.2 s.
In addition to the actuation uncertainties, we conducted the second test in a constrained environment with the soft water bag as depicted in Figure 5F.
In Figure 9B, the shape control error using the model-based method does not show regulation but demonstrate oscillations around a large value after the collision, while ours show the asymptotic convergences to zeros, which can be also indicated by the resulting CPDE.
Although it is difficult for a priori knowledge of the robot model to be accurate when considering uncertainties and perturbations, from the results with these unexpected conditions, we can confirm that the proposed learning-based
controller can online update the unknown model of continuum robot and robustly drive its 3-D shape towards the desired configuration.

\begin{figure}[!ht]
  \centering
  \includegraphics[width=0.95\linewidth]{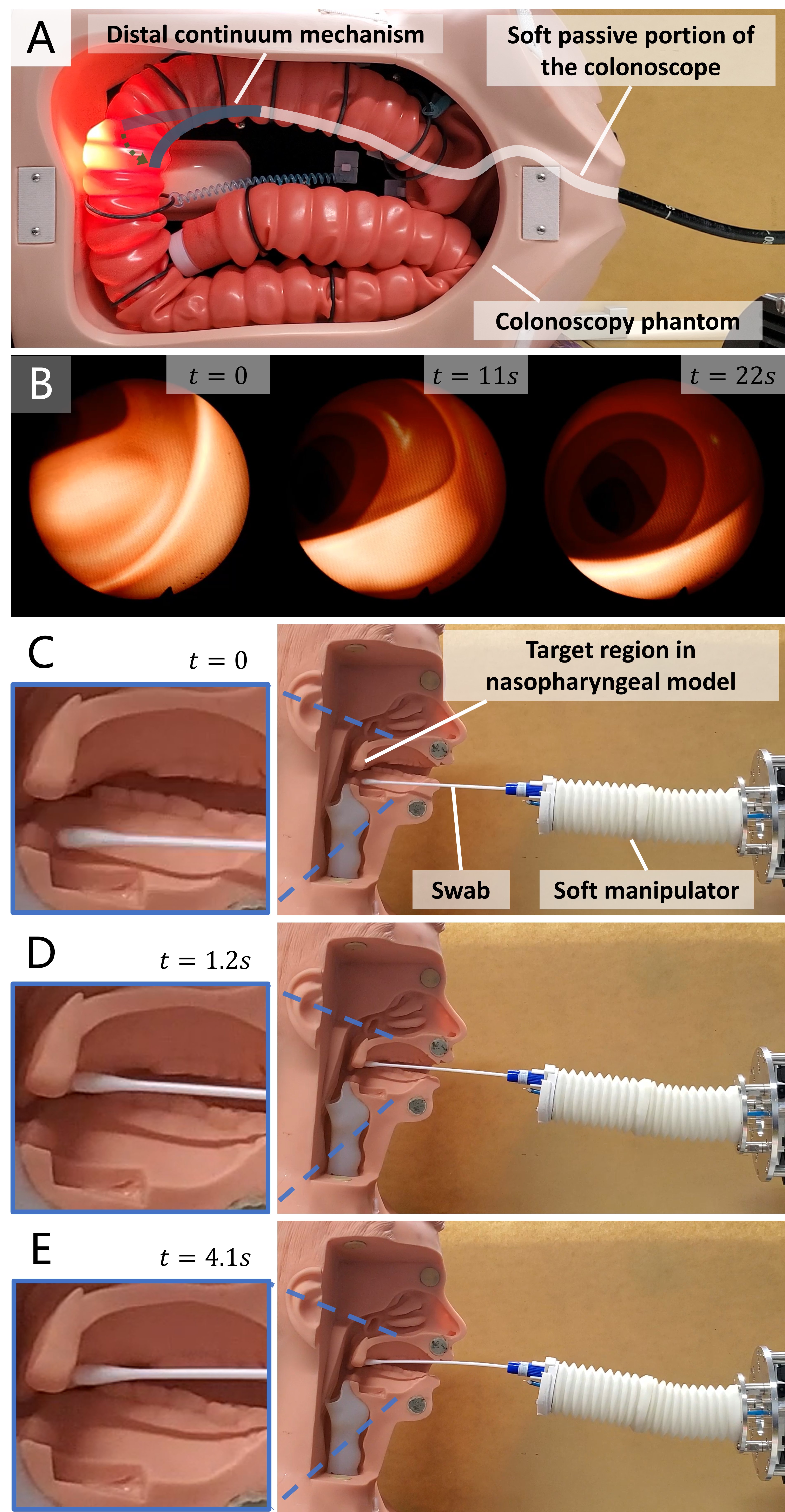}
  \caption{Task setups and snapshots of phantom experiments. Shape control for robotic colonoscopy: (A) setup and (B) endoscopic views. OP swab sampling: (C) initial setup, (D) interaction with target, and (E) shape control completion.}
  \label{fig:expt5}
\end{figure}

\subsection{Case Studies in Medical Contexts}
Two case studies were performed including shape control for robotic colonoscopy and oropharyngeal (OP) swab sampling.
We deployed RACS with 400 mm length inside a colonoscopy phantom (Kyoto Kagaku M40) for automatic shape control of its distal continuum mechanism.
The setup is shown in Figure 10A, where the distal part at the corner has to bend downwards to acquire a better endoscopic view for insertion guidance.
Figure 10B presents the snapshots of endoscopic view 
controlled from an unclear one to a feasible one for continuing insertion.
Note that RACS has a soft passive part with 280 mm length inside the phantom, which was not fixed and could collide with the phantom frequently resulting in unknown disturbances to the system.
Hence, the results behave some oscillations in Figure 9C, while their convergences show that our algorithm can adaptively update the time-varying model of continuum robot and simultaneously servo-control it to the desired shape against unknown disturbances.

The setup of OP sampling is illustrated in Figure 10C, where a swab is attached to the distal endpoint of SCM and the swab tip has to interact with the target region in a nasopharyngeal model (KJR/7208) for successful sampling.
The swab is deformed by the interaction as shown in Figure 10D and 10E, and the robot model during the interaction is regarded as different from the initial one.
Figure 9D shows that the 
endpoint position error $\| \Delta \bm{r}_{1} \|$ behaves slower convergence after the collision 
with the target, yet the algorithm adaptively keeps the error converging to zero with approaching the desired configuration.
These results of phantom experiments support the claim that our controller can perform robust compliance to the unstructured environments, online learning of the time-varying model, and simultaneously servo-control the 3-D shape of continuum robot 
in these credible contexts.

\subsection{Discussion}
Apart from CPDE and shape errors presented above, the rank of the matrix $\widehat{\bm{J}}_c (\bm{q}) \widehat{\bm{J}}_c^{+} (\bm{q}) \in \mathbb{R}^{m \times m}$ 
was monitored to prevent its reduction weakening the robot's manipulability.
The control inputs $\dot{\bm{q}}$ were recorded as well and did not reach the threshold, thus avoiding fast motion of the robot.
From all these results using two different continuum robots, we can observe that our control method can adaptively control the 3-D shape of the continuum robot without any priori model in various unstructured environments.
As for SCM featured with variable length, DEP-BTA composed of endpoint positioning as well as bending/twisting of each soft section, is better to describe and control its shape, rather than the feature of two points which may bring unexpected extension/compression to the soft sections.
The limitation of the approach is that the deployed shape sensing does not consider the variable arc length of SCM, though we can approximate and control its time-varying length by positioning its distal endpoint.
We will enable an accurate length estimation using FBG sensors to cope with this issue.
3-D shape control of continuum robot with more sections will also be investigated. 

\section{Conclusion}
In this work, we propose a novel and generic data-driven approach for adaptive 3-D shape control of continuum and soft robots using FBG shape feedback in unstructured environments.
An enhanced multi-core FBG-based method with an accurate shape sensing algorithm is employed to obtain the robot shape without external sensing units;
thus, it can be utilized in MIS with occlusions of sight-of-view.
The proposed algorithm enables robust model learning 
and simultaneous servo-control of the robot imposing the desired shape without any priori knowledge,
thereby facilitating its applications with highly nonlinear uncertainties and unknown interactions.
A learning-based shape flow predictor together with a composite adaptation law are introduced for asymptotic regulations of shape estimation errors and online update of learning parameters.
The stability of the close-loop system under our 
controller has been theoretically proved by Lyapunov theory.

To evaluate the performance of the proposed method,
detailed experiments have been conducted using two different continuum robots, including RACS and SCM, both equipped with multi-core FBG fibers for 3-D shape sensing.
It is worth noting that SCM is featured with variable length and stiffness.
The results that consist of the converging paths of the distal endpoint and the asymptotic regulation of the shape control and estimation errors to zeros, demonstrate the versatility, adaptability, and superiority of our method in various unstructured environments and phantom experiments against unexpected 
disturbances, which facilitate its practical deployments to various continuum robots and flexible instruments.

Although we evaluate the performance of our method in 3-D space, we only test it with desired configurations in C-shape, and those described in S-shape 
will be investigated 
in the future.
Since the variable length of SCM would affect the performance, a key issue is to acquire the accurate sensing length of the robot \cite{wang2016visual}.
For this purpose, an assisting device and methodology using FBG sensors will be designed.
Moreover, due to the versatility of the proposed controller, it can be extended for more practical tasks, e.g., soft tissue manipulation \cite{navarro2016automatic}, and eye-in-hand visual servoing \cite{fang2019vision, wang2020eye, fang2021soft}.
By utilizing FBG data, the method will also be introduced in 
energy-based planning 
for underactuated control of continuum and soft robots within efficient and safe frameworks.

\appendices
\section{3-D shape servoing algorithm}
\let\oldnl\nl
\newcommand{\nonl}{\renewcommand{\nl}{\let\nl\oldnl}}
\begin{algorithm}[ht]
  \small
  \DontPrintSemicolon
  \nonl\textbf{Initialization}:\;
  $\bm{x}_d \longleftarrow$ define a shape description and a desired shape\;
  $\epsilon_e \longleftarrow$ specify a positive scalar as threshold\;
  $\widehat{\bm{W}}_i, i \in \{1, 2, ..., m\} \longleftarrow$ initialize NN weight matrices\;
  \nonl\textbf{Shape servoing}:\;
  \While{\textup{shape error norm} $\| \bm{e} \| > \epsilon_e$}{
    $\dot{\widehat{\bm{x}}}, \dot{\widetilde{\bm{x}}} \longleftarrow$ predict shape flow and calculate its error\;
    $\widetilde{\bm{x}} \longleftarrow$ compute shape estimation error\;
    $\bm{q} \longleftarrow$ measure actuator position\;
    $\widehat{\overline{\bm{W}}} \longleftarrow$ update vector of NN weights\;
    $\widehat{\bm{J}}_c (\bm{q}) \longleftarrow$ compute combined Jacobian matrix\;
    $\dot{\bm{q}} \longleftarrow$ compute velocity controller\;
    Command actuator motion $\dot{\bm{q}}$\;
    $\bm{r}, \bm{x} \longleftarrow$ generate 3-D shape from FBG-measured points\;
    $\bm{e} \longleftarrow$ calculate shape control error\;
  }
  \caption{3-D shape servoing algorithm}
\end{algorithm}

\section{Proof of the closed-loop system stability}
Define a Lyapunov candidate function as
\begin{equation}
\begin{aligned}
    \label{eq:lyapunov}
    \mathcal{V} =
	\frac{1}{2} k_e \bm{e}^{\intercal} \bm{e}
	+ \frac{1}{2} k_x \widetilde{\bm{x}}^{\intercal} \widetilde{\bm{x}}
	+ \frac{1}{2} \widetilde{\overline{\bm{W}}}^{\intercal} \bm{\Gamma}_W \widetilde{\overline{\bm{W}}}
	+ k_r \mathcal{R}
\end{aligned}
\end{equation}
with the auxiliary function $\mathcal{R} \left( t \right)$ defined as
\begin{equation}
\begin{aligned}
    \label{eq:auxiliary_R}
    \mathcal{R} \left( t \right) =
    \beta_x \sum_{i=1}^m \left| \widetilde{\bm{x}}_i\left( 0 \right) \right|
    - \widetilde{\bm{x}}^{\intercal} \left( 0 \right) \bm{\delta} \left( 0 \right)
	- \mathcal{H} \left( t \right)
	+ \mathcal{H} \left( 0 \right)
\end{aligned}
\end{equation}
where $i = 1, 2, \cdots, m$ represents the $i$-th element of $\widetilde{\bm{x}}$, and the auxiliary function $\mathcal{H} \left( t \right)$ \cite{dinh2010dynamic} is generated by
\begin{equation}
\begin{aligned}
    \dot{\mathcal{H}}
    = \bm{r}_x^{\intercal} 
    \left(
    \bm{\delta} 
    - \beta_x {\rm sat} \left( \widetilde{\bm{x}} \right)
    \right)
\end{aligned}
\end{equation}

The time derivative of $\mathcal{R} \left( t \right)$ is derived as
\begin{equation}
\begin{aligned}
    \label{eq:psdfunction}
    \dot{\mathcal{R}}
    = - \dot{\mathcal{H}}
    = - \bm{r}_x^{\intercal} 
    \left(
    \bm{\delta} 
    - \beta_x {\rm sat} \left( \widetilde{\bm{x}} \right)
    \right)
\end{aligned}
\end{equation}
and $\mathcal{R} \left( t \right)$ is positive semi-definite if the sufficient condition 
\begin{equation}
\begin{aligned}
    \label{eq:betax}
    \beta_x \ge b_{\delta 1} + \frac{b_{\delta 2}}{\alpha_x}
\end{aligned}
\end{equation}
in Equation (\ref{eq:sufficient}) is satisfied, which is proved in Appendix C.

Differentiating $\mathcal{V}$ in Equation (\ref{eq:lyapunov}) and substituting Equations (\ref{eq:modeling}), (\ref{eq:estimator}), (\ref{eq:adaptive}), and (\ref{eq:psdfunction}) results in
\begin{equation}
\begin{aligned}
    \label{eq:vdot1}
	\dot{\mathcal{V}}
	& =
	- k_x \alpha_x \widetilde{\bm{x}}^{\intercal} \widetilde{\bm{x}}
    + k_x \widetilde{\bm{x}}^{\intercal}
    \left(
    \bm{\delta}
    - \beta_x {\rm sat}(\widetilde{\bm{x}})
    \right)
    + k_e \bm{e}^{\intercal} \bm{J}(\bm{q}) \dot{\bm{q}} 
	\\
	& \quad
	+ k_e \bm{e}^{\intercal} \bm{d}
	- k_e \widetilde{\overline{\bm{W}}}^{\intercal}
	\bm{Q} (\dot{\bm{q}}) \bm{\Theta} (\bm{q}) \bm{e}
	- k_r \widetilde{\overline{\bm{W}}}^{\intercal}
	\bm{Q} (\dot{\bm{q}}) \bm{\Theta} (\bm{q}) \bm{r}_x
    \\
	& \quad
	- k_r \bm{r}_x^{\intercal} 
    \left(
    \bm{\delta} 
    - \beta_x {\rm sat} \left( \widetilde{\bm{x}} \right) 
    \right)
\end{aligned}
\end{equation}

Substituting Equations (\ref{eq:JJW}) and (\ref{eq:rxW}), we have
\begin{equation}
\begin{aligned}
	\dot{\mathcal{V}}
	& =
	- k_x \alpha_x \widetilde{\bm{x}}^{\intercal} \widetilde{\bm{x}}
    + k_x \widetilde{\bm{x}}^{\intercal}
    \left(
    \bm{\delta}
    - \beta_x {\rm sat}(\widetilde{\bm{x}})
    \right)
    + k_e \bm{e}^{\intercal} \widehat{\bm{J}}_c (\bm{q}) \dot{\bm{q}}
	\\
	& \quad
	+ k_e \bm{e}^{\intercal} \bm{\delta}
	- k_r \bm{r}_x^{\intercal} \bm{r}_x
\end{aligned}
\end{equation}

Using the controller in Equation (\ref{eq:control}) and taking the upper bound with Equation (\ref{eq:disbound}), we can further have
\begin{equation}
\begin{aligned}
	\dot{\mathcal{V}}
    & =
	- k_x \alpha_x \widetilde{\bm{x}}^{\intercal} \widetilde{\bm{x}}
    + k_x \widetilde{\bm{x}}^{\intercal}
    \left(
    \bm{\delta}
    - \beta_x {\rm sat}(\widetilde{\bm{x}})
    \right)
	\\
	& \quad
	+ k_e \bm{e}^{\intercal} \bm{\delta}
	- k_e \bm{e}^{\intercal}
	\left(
	k_c \bm{e} + k_s {\rm sat}(\bm{e})
    \right)
	- k_r \bm{r}_x^{\intercal} \bm{r}_x
	\\
	& \le
	- k_x \alpha_x \widetilde{\bm{x}}^{\intercal} \widetilde{\bm{x}}
	- k_x \left( \beta_x - b_{\delta 1} \right)
	\left\| \widetilde{\bm{x}} \right\|
	- k_e k_c \bm{e}^{\intercal} \bm{e}
	\\
	& \quad
	- k_e \left( k_s - b_{\delta 1} \right)
	\left\| \bm{e} \right\|
	- k_r \bm{r}_x^{\intercal} \bm{r}_x
\end{aligned}
\end{equation}

From Equation (\ref{eq:betax}), we can conclude that $- k_x \left( \beta_x - b_{\delta 1} \right) \left\| \widetilde{\bm{x}} \right\| \le 0$ since the condition is satisfied
\begin{equation}
\begin{aligned}
    \beta_x \ge b_{\delta 1} + \frac{b_{\delta 2}}{\alpha_x}
    \ge b_{\delta 1}
\end{aligned}
\end{equation}

If the controller parameter $k_s$ is chosen sufficiently large such that $ k_s \ge b_{\delta 1} $ in Equation (\ref{eq:sufficient}) is satisfied, we have $- k_e \left( k_s - b_{\delta 1} \right) \left\| \bm{e} \right\| \le 0$ and obtain that
\begin{equation}
\begin{aligned}
	\dot{\mathcal{V}}
    \le
	- k_x \alpha_x \widetilde{\bm{x}}^{\intercal} \widetilde{\bm{x}}
	- k_e k_c \bm{e}^{\intercal} \bm{e}
	- k_r \bm{r}_x^{\intercal} \bm{r}_x
	\le 0
\end{aligned}
\end{equation}

Since $\mathcal{V} >0$ and $\dot{\mathcal{V}} \le 0$, $V$ is bounded, which directly implies that the shape control error $\bm{e}$, shape estimation error $\widetilde{\bm{x}}$, and estimation error of NN weights $\widetilde{\overline{\bm{W}}}$ are all bounded.
The boundedness of $\dot{\bm{q}}$, and that of activation functions $\bm{\theta}_i (\bm{q}), i \in \{1, 2, ..., m\}$ as well as their derivatives w.r.t. the corresponding arguments \cite{lewis2002neuro} guarantee the boundedness of $\bm{\Theta}$ and $\dot{\bm{\Theta}}$ from Equation (\ref{eq:block_theta}).
Because $\bm{Q}$ and $\bm{\delta}$ are bounded by Equations (\ref{eq:block_q}) and (\ref{eq:disbound}), respectively, their boundedness together with those of $\bm{\Theta}$, $\widetilde{\overline{\bm{W}}}$, $\widetilde{\bm{x}}$, and ${\rm sat} \left( \cdot \right)$ function ensure that the shape flow error $\dot{\widetilde{\bm{x}}}$ is bounded from Equation (\ref{eq:estimator}).
From Equation (\ref{eq:rxW}), the boundedness of $\widetilde{\bm{x}}$ and $\dot{\widetilde{\bm{x}}}$ implies the boundedness of the filtered shape estimation error $\bm{r}_x$.
The $\rm{proj} \left( \cdot \right)$ function in Equation (\ref{eq:adaptive}) indicates the boundedness of $\widehat{\overline{\bm{W}}}$, $\dot{\widehat{\overline{\bm{W}}}}$, and $\dot{\widetilde{\overline{\bm{W}}}}$.
Since $\dot{\bm{q}}$ and $\ddot{\bm{q}}$ are bounded,
and so are $\bm{W}_i$, $\bm{\theta}_i (\bm{q}), i \in \{1, 2, ..., m\}$, and $\bm{E}_J$ \cite{lewis2002neuro}, $\bm{J}_c (\bm{q})$ and $\dot{\bm{Q}}$ are bounded from Equations (\ref{eq:Jnn}) and (\ref{eq:block_q}), respectively.
From Equation (\ref{eq:modeling}), the boundedness of $\dot{\bm{q}}$, $\bm{J}_c (\bm{q})$, and $\bm{d}$ also indicates that the shape feature is varied continuously, i.e., $\dot{\bm{x}}$ is bounded and so is $\dot{\bm{e}}$.
Since $\bm{\Theta}$, $\dot{\bm{\Theta}}$, $\bm{Q}$, $\dot{\bm{Q}}$, $\widetilde{\overline{\bm{W}}}$, $\dot{\widetilde{\overline{\bm{W}}}}$, $\dot{\bm{\delta}}$ from Equation (\ref{eq:disbound}), ${\rm sat} \left( \cdot \right)$ function with its derivative w.r.t. $\widetilde{\bm{x}}$, and $\dot{\widetilde{\bm{x}}}$ are all bounded, we can conclude that $\dot{\bm{r}}_x$ is bounded from Equation (\ref{eq:rxW}).
Therefore, $\bm{e}$, $\widetilde{\bm{x}}$, and $\bm{r}_x$ are uniformly continuous.
By Barbalat's lemma, we can prove that
\begin{equation}
\begin{aligned}
    &
    \lim\limits_{t \to \infty} \bm{e} = \bm{0}, 
    \quad
    \lim\limits_{t \to \infty} \bm{r}_x = \bm{0},
    \quad
    \lim\limits_{t \to \infty} \widetilde{\bm{x}} = \bm{0}
\end{aligned}
\end{equation}
As $t \rightarrow \infty$, since $\bm{r}_x \rightarrow \bm{0}$ and $\widetilde{\bm{x}} \rightarrow \bm{0}$, we have $\dot{\widetilde{\bm{x}}} \rightarrow \bm{0}$ following Equation (\ref{eq:rxW}).
From Equation (\ref{eq:estimator}), we can prove the asymptotic minimization of the estimation error of NN weights $\widetilde{\overline{\bm{W}}}$.
Consequently, the shape control error, shape estimation error, and shape flow error are all convergent to zeros,
and the online learning of the combined Jacobian $\bm{J}_c (\bm{q})$ using our adaptive NNs approximation is guaranteed.

\section{Proof of positive semi-definiteness of $\mathcal{R} \left( t \right)$}
Recall that the auxiliary function $\mathcal{H} \left( t \right)$ is generated by
\begin{equation}
\setlength{\abovedisplayskip}{3pt}
\setlength{\belowdisplayskip}{3pt}
    \dot{\mathcal{H}}
    = \bm{r}_x^{\intercal} 
    \left(
    \bm{\delta} 
    - \beta_x {\rm sat} \left( \widetilde{\bm{x}} \right)
    \right)
\end{equation}

By integrating $\dot{\mathcal{H}}$ and substituting $\bm{r}_x = \dot{\widetilde{\bm{x}}} + \alpha_x \widetilde{\bm{x}}$ in Equation (\ref{eq:rxW}), it is obtained that
\begin{equation}
\setlength{\abovedisplayskip}{3pt}
\setlength{\belowdisplayskip}{3pt}
\begin{aligned}[b]
    \mathcal{H} \left( t \right) 
    & = 
    \mathcal{H} \left( 0 \right) 
    + \int_{0}^{t}
    \left( 
    \dot{\widetilde{\bm{x}}} + \alpha_x \widetilde{\bm{x}} 
    \right)^{\intercal} 
    \left(
    \bm{\delta} - \beta_x {\rm sat} \left( \widetilde{\bm{x}} \right) 
    \right)
    d \tau
     \\
    & = 
    \mathcal{H} \left( 0 \right) 
    + \int_{0}^{t}
    \dot{\widetilde{\bm{x}}}^{\intercal} \bm{\delta}
    d \tau
    - \int_{0}^{t}
    \beta_x \dot{\widetilde{\bm{x}}}^{\intercal} 
    {\rm sat} \left( \widetilde{\bm{x}} \right) 
    d \tau
     \\
    & \quad
    + \int_{0}^{t}
    \alpha_x \widetilde{\bm{x}}^{\intercal} 
    \left(
    \bm{\delta} - \beta_x {\rm sat} \left( \widetilde{\bm{x}} \right) 
    \right)
    d \tau
     \\
    & = 
    \mathcal{H} \left( 0 \right) 
    + \widetilde{\bm{x}}^{\intercal} \bm{\delta}
    - \widetilde{\bm{x}}^{\intercal} \left( 0 \right) 
    \bm{\delta} \left( 0 \right) 
    - \int_{0}^{t}
    \widetilde{\bm{x}}^{\intercal} \dot{\bm{\delta}}
    d \tau
     \\
    & \quad
    - \beta_x \sum_{i=1}^m 
    \left| \widetilde{\bm{x}}_i\left( t \right) \right|
    + \beta_x \sum_{i=1}^m 
    \left| \widetilde{\bm{x}}_i\left( 0 \right) \right|
     \\
    & \quad
    + \int_{0}^{t}
    \alpha_x \widetilde{\bm{x}}^{\intercal} 
    \left(
    \bm{\delta} - \beta_x {\rm sat} \left( \widetilde{\bm{x}} \right) 
    \right)
    d \tau
\end{aligned}
\end{equation}

Since $\sum_{i=1}^m \left| \widetilde{\bm{x}}_i\left( t \right) \right| \ge \left\| \widetilde{\bm{x}} \right\|$, taking the upper bound of above equation yields
\begin{equation}
\setlength{\abovedisplayskip}{3pt}
\setlength{\belowdisplayskip}{3pt}
\begin{aligned}
    \mathcal{H} \left( t \right) 
    & \le
    \mathcal{H} \left( 0 \right) 
    + \left\| \widetilde{\bm{x}} \right\| 
    \left\| \bm{\delta} \right\|
    - \widetilde{\bm{x}}^{\intercal} \left( 0 \right) 
    \bm{\delta} \left( 0 \right) 
    \\
    & \quad
    + \int_{0}^{t}
    \left\| \widetilde{\bm{x}} \right\|
    \| \dot{\bm{\delta}} \|
    d \tau
    - \beta_x \left\| \widetilde{\bm{x}} \right\|
    + \beta_x \sum_{i=1}^m 
    \left| \widetilde{\bm{x}}_i\left( 0 \right) \right|
    \\
    & \quad
    + \int_{0}^{t}
    \alpha_x \left\| \widetilde{\bm{x}} \right\|
    \left\| 
    \bm{\delta} - \beta_x {\rm sat} \left( \widetilde{\bm{x}} \right) 
    \right\|
    d \tau
\end{aligned}
\end{equation}

Substituting the upper bounds of $\bm{\delta}$ and $\dot{\bm{\delta}}$ from Equation (\ref{eq:disbound}), we can obtain that
\begin{equation}
\setlength{\abovedisplayskip}{3pt}
\setlength{\belowdisplayskip}{3pt}
\begin{aligned}
    \mathcal{H} \left( t \right) 
    & \le
    \mathcal{H} \left( 0 \right)
    - \widetilde{\bm{x}}^{\intercal} \left( 0 \right) 
    \bm{\delta} \left( 0 \right) 
    + \beta_x \sum_{i=1}^m 
    \left| \widetilde{\bm{x}}_i\left( 0 \right) \right|
    \\
    & \quad
    - \left( \beta_x - b_{\delta 1} \right) 
    \left\| \widetilde{\bm{x}} \right\|
    \\
    & \quad
    - \int_{0}^{t}
    \alpha_x \left( 
    \beta_x - b_{\delta 1} - \frac{b_{\delta 2}}{\alpha_x} \right)
    \left\| \widetilde{\bm{x}} \right\|
    d \tau
\end{aligned}
\end{equation}

We can conclude that the last two terms on the right of the above inequality both negative semi-definite if the condition 
\begin{equation}
\setlength{\abovedisplayskip}{3pt}
\setlength{\belowdisplayskip}{3pt}
    \beta_x \ge b_{\delta 1} + \frac{b_{\delta 2}}{\alpha_x}
\end{equation}
in Equation (\ref{eq:sufficient}) is satisfied, then we further have
\begin{equation}
\setlength{\abovedisplayskip}{3pt}
\setlength{\belowdisplayskip}{3pt}
    \mathcal{H} \left( t \right) 
    \le
    \mathcal{H} \left( 0 \right)
    - \widetilde{\bm{x}}^{\intercal} \left( 0 \right) 
    \bm{\delta} \left( 0 \right) 
    + \beta_x \sum_{i=1}^m 
    \left| \widetilde{\bm{x}}_i\left( 0 \right) \right|
\end{equation}
and prove $\mathcal{R} \left( t \right) \ge 0$ in Equation (\ref{eq:auxiliary_R}).




\ifCLASSOPTIONcaptionsoff
  \newpage
\fi



%

\bibliographystyle{./bib/IEEEtran}
\bibliography{./bib/references}

%








\end{document}